%% file: main.tex
\documentclass[english,ruled]{article}
\usepackage[T1]{fontenc}
\usepackage[latin9]{inputenc}
\usepackage{verbatim}
\usepackage{subcaption}
\usepackage{algorithm2e}
\usepackage{amsmath}
\usepackage{amsthm}
\usepackage{etoolbox}
\usepackage{amssymb}
\usepackage{wasysym}
\usepackage{graphicx}
\usepackage{xcolor}
\usepackage{mathtools}
\usepackage{multicol}
\usepackage{multirow}
\usepackage{geometry}
\usepackage{booktabs}
\usepackage{enumitem}
\usepackage{setspace}
\makeatletter
\usepackage[toc,page,header]{appendix}
\usepackage{minitoc}
\usepackage{ifthen}
\newboolean{doublecolumn}
\setboolean{doublecolumn}{false}
\newboolean{algolink}
\setboolean{algolink}{true}
\newboolean{arxiv}
\setboolean{arxiv}{true}
\usepackage{pgfplots}
\usepackage{pdflscape}
\usepackage[pagebackref=true]{hyperref}
\usepackage{cleveref}
\usepackage{authblk}
\usepackage{todonotes}
\pgfplotsset{compat=1.15}

\newboolean{proofinpaper}
\setboolean{proofinpaper}{false}

\newcommand{\inputinpaper}[1]{%
  \ifbool{proofinpaper}{\begin{proof}
						\input{#1}{}%
						\end{proof}}}

\newcommand{\inputinappendix}[1]{%
\ifbool{proofinpaper}{}{\input{#1}}%
}

\makeatletter
\renewcommand\paragraph{%
  \@startsection{paragraph}{4}{\z@}%
    {-1.5ex\@plus -0.5ex \@minus -0.2ex}% space before
    {-.5em \@plus -.1em}%               
    {\normalfont\normalsize\bfseries}%
}
\makeatother
\newcommand{\E}{\mathbb{E}}

\newcommand{\inner}[2]{\langle{#1},{#2}\rangle}
\newcommand{\assign}{\coloneqq}

\newcommand{\tmop}[1]{\ensuremath{\operatorname{#1}}}
\newcommand{\tmtextbf}[1]{\text{{\bfseries{#1}}}}

\newcommand{\tmem}[1]{\textit{#1}}
\newcommand{\mathd}{\mathrm{d}}
\newcommand{\mathe}{\mathrm{e}}
\definecolor{deepskyblue}{rgb}{0.0, 0.75, 1.0}

\input{macros.tex}

\renewcommand*\backref[1]{\ifx#1\relax \else (cited on #1) \fi}
\theoremstyle{plain}
\newtheorem{lem}{\protect\lemmaname}[section]
\theoremstyle{remark}
\newtheorem{rem}{\protect\remarkname}
\theoremstyle{plain}
\newtheorem{thm}{\protect\theoremname}[section]
\theoremstyle{plain}
\newtheorem{prop}{\protect\propositionname}[section]
\providecommand{\corollaryname}{Corollary}
\theoremstyle{plain}
\newtheorem{coro}{\protect\corollaryname}[section]
\theoremstyle{plain}
\newtheorem{exple}{\protect\examplename}[section]
\theoremstyle{plain}
\newtheorem{definition}{\protect\definitionname}[section]

\providecommand{\lemmaname}{Lemma}
\providecommand{\remarkname}{Remark}
\providecommand{\theoremname}{Theorem}
\providecommand{\examplename}{Example}
\providecommand{\propositionname}{Proposition}
\providecommand{\definitionname}{Definition}

\newcommand{\oco}{{\texttt{OCO}}}
\newcommand{\ogd}{{\texttt{OGD}}}
\newcommand{\adanorm}{{\texttt{AdaGrad-Norm}}}
\newcommand{\adaftrl}{{\texttt{AdaFTRL}}}
\newcommand{\adagrad}{{\texttt{AdaGrad}}}
\newcommand{\bgd}{{\texttt{BGD}}}

\crefdefaultlabelformat{#2\textbf{#1}#3} % <-- Only #1 in \textbf
\crefname{section}{\textbf{section}}{\textbf{sections}}
\Crefname{section}{\textbf{Section}}{\textbf{Sections}}
\crefname{thm}{\textbf{Theorem}}{\textbf{theorems}}
\Crefname{thm}{\textbf{Theorem}}{\textbf{Theorems}}
\crefname{lem}{\textbf{Lemma}}{\textbf{lemmas}}
\Crefname{lem}{\textbf{Lemma}}{\textbf{Lemmas}}
\crefname{prop}{\textbf{proposition}}{\textbf{propositions}}
\Crefname{prop}{\textbf{Proposition}}{\textbf{Propositions}}
\crefname{algorithm}{\textbf{algorithm}}{\textbf{algorithms}}
\Crefname{algorithm}{\textbf{Algorithm}}{\textbf{Algorithms}}
\crefname{coro}{\textbf{Corollary}}{\textbf{corollaries}}
\Crefname{coro}{\textbf{Corollary}}{\textbf{corollaries}}
\crefname{definition}{\textbf{Definition}}{\textbf{definitions}}
\Crefname{definition}{\textbf{Definition}}{\textbf{definitions}}
\crefname{table}{\textbf{Table}}{\textbf{tables}}
\Crefname{table}{\textbf{Table}}{\textbf{tables}}
\crefname{figure}{\textbf{Figure}}{\textbf{figures}}
\Crefname{figure}{\textbf{Figure}}{\textbf{figures}}
\crefname{exple}{\textbf{Example}}{\textbf{examples}}
\Crefname{exple}{\textbf{Example}}{\textbf{examples}}

\renewcommand{\E}{\mathbb{E}}

\begin{document}

\title{Small Gradient Norm Regret for Online Convex Optimization}

\author[1]{Wenzhi Gao\thanks{gwz@stanford.edu}}
\author[2]{Chang He\thanks{hec@umn.edu}}
\author[1,3]{Madeleine Udell\thanks{udell@stanford.edu}}
\affil[1]{ICME, Stanford University}
\affil[2]{Department of Industrial and Systems Engineering, University of Minnesota}
\affil[3]{Department of Management Science and Engineering, Stanford University}

\maketitle

\begin{abstract}
This paper introduces a new problem-dependent regret measure for online convex optimization with smooth losses. The notion, which we call the $G^\star$ regret, depends on the cumulative squared gradient norm evaluated at the decision in hindsight $\sum_{t=1}^T \|\nabla \ell(x^\star)\|^2$. We show that the $G^\star$ regret strictly refines the existing $L^\star$ (small loss) regret, and that it can be arbitrarily sharper when the losses have vanishing curvature around the hindsight decision.
 We establish upper and lower bounds on the $G^\star$ regret and extend our results to dynamic regret and bandit settings. As a byproduct, we refine the existing convergence analysis of stochastic optimization algorithms in the interpolation regime.  Some experiments validate our theoretical findings.
\end{abstract}

\input{sec_intro.tex}

\input{sec_gstar.tex}
\input{sec_extension.tex}

\input{sec_application.tex}
\input{sec_exp.tex}

\input{sec_conclusion.tex}

\section*{Acknowledgement}
We appreciate the constructive feedback from Professor Qi Deng from Shanghai Jiao Tong University.

\renewcommand \thepart{}
\renewcommand \partname{}

\bibliography{ref.bib}
\bibliographystyle{plain}

\doparttoc
\faketableofcontents
\part{}

\newpage
\appendix

\addcontentsline{toc}{section}{Appendix}
\part{Appendix} 
\parttoc

\newpage
\input{sec_app.tex}

\end{document}

%% file: macros.tex
\global\long\def\inner#1#2{\langle#1,#2\rangle}%

\global\long\def\vertiii#1{\left\vert \kern-0.25ex  \left\vert \kern-0.25ex  \left\vert #1\right\vert \kern-0.25ex  \right\vert \kern-0.25ex  \right\vert }%

\global\long\def\diam{\mathrm{diam}}%

\global\long\def\argmin{\operatornamewithlimits{arg\,min}}%

\global\long\def\dom{\mathrm{dom}}%

\global\long\def\and{\mathrm{and}}%

\global\long\def\dist{\mathrm{dist}}%

\global\long\def\dom{\operatornamewithlimits{dom}}%

\global\long\def\Ebb{\mathbb{E}}%

\global\long\def\Rbb{\mathbb{R}}%

\global\long\def\Ocal{\mathcal{O}}%

\global\long\def\Pcal{\mathcal{P}}%

\global\long\def\Sbb{\mathbb{S}}%

\global\long\def\Xcal{\mathcal{X}}%

%% file: sec_intro.tex
\section{Introduction} \label{sec:intro}
Online convex optimization (\oco) is a powerful framework for
handling sequential decision-making problems
{\cite{hazan2016introduction,orabona2019modern}}. In an
{\oco} problem, a learner makes sequential decisions over time
horizon $T$ in an adversarial setting: at each time step $t = 1, \ldots, T$, the
learner chooses some decision $x^t$ from a convex set $\mathcal{X} \subseteq
\mathbb{R}^n$ and receives a convex loss $\ell_t : \mathbb{R}^n
\rightarrow \mathbb{R}$ specified by the adversary. The
goal of an {\oco} algorithm is to make decisions $\{ x^t \}_{t =
1}^T$ that achieve low cumulative regret with respect to an arbitrary decision
$x \in \mathcal{X}$
\[ 
\textstyle \rho_T (x) \assign \sum_{t = 1}^T \ell_t (x^t) - \sum_{t = 1}^T \ell_t (x),
\]
and in particular, the best decision in hindsight $x^{\star} \in \argmin_{x \in
\mathcal{X}}  \sum_{t = 1}^T \ell_t (x)$. 
 Despite the powerful adversary, existing {\oco} algorithms can achieve sublinear regret under different assumptions on the losses $\ell_t$.
For example, when the losses are convex and Lipschitz,
{\cite{zinkevich2003online}} shows that online gradient descent
({\ogd}) achieves $\mathcal{O} ( \sqrt{T} )$ regret; if
$\ell_t$ are additionally strongly convex or exponentially concave, then
$\mathcal{O} (\log T)$ is attainable using {\ogd} or online Newton
step {\cite{hazan2007logarithmic}}. These regret bounds are minimax optimal in the corresponding loss function classes
{\cite{abernethy2008optimal}}.\\

Beyond the assumptions discussed above, another popular condition on the loss functions is
smoothness, which is often satisfied by loss functions in machine learning
tasks {\cite{srebro2010smoothness}}. When the loss functions are non-negative and have Lipschitz
continuous gradients, it is possible to obtain a problem-dependent regret $\mathcal{O} (
\sqrt{L_T^{\star}} )$ {\cite{srebro2010smoothness}}, where $L_T^{\star}
= \min_{x \in \mathcal{X}}  \sum_{t = 1}^T \ell_t (x) = \sum_{t = 1}^T \ell_t
(x^{\star})$ denotes the cumulative loss under the best decision in hindsight. This regret, often
known as the $L^{\star}$-regret or small-loss bound
{\cite{orabona2019modern,zhao2020dynamic,wang2020adapting,zhang2019adaptive}},
can be significantly sharper than the standard $\mathcal{O} ( \sqrt{T} )$
regret upper bound if $L_T^{\star} =\mathcal{O} (1)$, while still maintaining the worst-case $\mathcal{O} ( \sqrt{T} )$ regret. Recent work has developed parameter-free online learning algorithms that achieve
$\mathcal{O} ( \sqrt{L_T^{\star}} )$ regret
{\cite{zhao2020dynamic,zhao2024adaptivity}} as well as guarantees in terms of other problem-dependent
quantities.\\

Despite its strengths, the small loss regret ($L^{\star}$ regret) has certain limitations. First, most existing results for $L^{\star}$ regret assume that the losses are non-negative. Although this assumption can be removed by
recentering each loss via
$\bar{\ell}_t(x) \assign \ell_t(x) - \inf_{x \in \Rbb^n} \ell_t(x)$, it implicitly requires the losses to be lower bounded, a condition that is violated even by linear losses. Moreover, the lack of translation invariance (under the addition of a constant) of $L_T^{\star}$ remains conceptually unsatisfactory. These limitations motivate the search for an alternative regret measure that avoids
these shortcomings.

\paragraph{Contributions.}
This paper addresses the above problems by introducing the
$G^{\star}$ regret, which depends on the quantity
$G_T^\star \assign\sum_{t = 1}^T \| \nabla \ell_t (x^\star) \|^2$,  the cumulative non-stationarity at the best decision in hindsight. In particular,
\begin{itemize}[leftmargin=10pt,itemsep=0pt]
  \item We show that the $G^{\star}$ regret is at least as tight as the $L^{\star}$ regret and overcomes all its drawbacks: it is translation invariant and does not require the losses to be lower bounded over $\Rbb^n$. Moreover, we prove that the $G^{\star}$ regret can be arbitrarily sharper than the $L^{\star}$ regret when the losses have a small local smoothness constant around $x^\star$. We support these claims with numerical experiments.
  
  \item We prove that existing algorithms that adapt to the $L^{\star}$ regret
  also adapt to the $G^{\star}$ regret, achieving an
  $\mathcal{O}(\sqrt{G_T^\star})$ bound with a matching lower bound.
  We further extend our analysis to dynamic regret and bandit settings, and showcase the implications of $G^{\star}$ regret in the context of stochastic optimization. We also show how to design smoothed versions of the loss functions to allow adaptation to the constrained small-loss bound.
\end{itemize} 

\subsection{Related literature}

We survey two existing problem-dependent regret bounds in {\oco} based on smoothness. For a comprehensive review of the problem-dependent regret bounds, see \cite{zhao2024adaptivity} and the references therein.

\paragraph{$L^{\star}$ regret.} The $L^{\star}$ regret, also known as the small
loss bound, can provide a tighter regret upper bound when the decision in hindsight
achieves a small cumulative loss.
\begin{equation} \label{eqn:Lstar-regret}
\textstyle L_T^{\star} \assign  \min_{x \in \mathcal{X}} \sum_{t = 1}^T [\ell_t (x) -
 \inf_{u \in \Rbb^n} \ell_t (u) ] = \sum_{t = 1}^T [\ell_t (x^\star) -
 \inf_{u \in \Rbb^n} \ell_t (u) ],
\end{equation}
where $x^\star \in \argmin_{x \in \Xcal} \sum_{t=1}^T \ell_t (x)$. In the literature, it is often assumed that $\{ \ell_t \}_{t=1}^T$ are non-negative, and an upper bound $\min_{x \in \mathcal{X}} \sum_{t = 1}^T
\ell_t (x)$, is typically used. For ease of later comparison, this paper adopts the definition \eqref{eqn:Lstar-regret}, the recentered version of the $L^{\star}$ regret. Existing {\oco} algorithms
can achieve $\mathcal{O} ( \sqrt{L^{\star}_T} )$ regret for general
smooth convex functions and $\mathcal{O} (\log L^{\star}_T)$ regret when the loss is exponentially concave or strongly convex {\cite{wang2020adapting,srebro2010smoothness}}. Recently, the $L^\star$ regret has also been extended to the adaptive regret setting \cite{yang2024small,zhang2019adaptive}.

\paragraph{Gradient variation bound.}Aside from the $L^{\star}$ regret, the
following gradient variation bound \cite{chiang2012online, mohri2016accelerating} is another widely adopted problem-dependent quantity. It often appears in the context of online learning with prediction, and depends on the maximum
cumulative variation of the gradients over $\mathcal{X}$
\begin{equation} \label{eqn:gradvar-regret}
	\textstyle V_T \assign    \sum_{t = 2}^T \max_{x \in \mathcal{X}} \| \nabla \ell_t (x) -
   \nabla \ell_{t - 1} (x) \|^2 .
\end{equation}
Existing {\oco} algorithms can achieve $\Ocal(\sqrt{V_T})$ regret \cite{chiang2012online,zhao2020dynamic,zhao2024adaptivity}, and recent work has extended these results beyond the standard Lipschitz continuous gradient setting \cite{xie2024gradient,zhao2025gradient}. 

\subsection{Notations}
Throughout the paper, we use $\| \cdot \|$ to denote the Euclidean norm and
$\langle \cdot, \cdot \rangle$ to denote the Euclidean inner product.
Given a closed convex set $\mathcal{X}$, $\Pi_{\mathcal{X}} [\cdot]$ denotes
the orthogonal projection onto $\mathcal{X}$. The diameter of a set is defined by $\diam (\mathcal{X}) \assign \sup_{x,y} \|x-y\|$. We use $\Sbb_n \assign \{x \in \Rbb^n: \|x\| = 1\}$ to denote  the unit sphere centered at the origin. For brevity, we use $\inf \ell \assign \inf_{x \in \Rbb^n} \ell(x)$ to denote the infimum of $\ell$ over $\Rbb^n$. We define $[n] \assign \{1, \ldots, n\}$.

%% file: sec_gstar.tex
\section{Small gradient norm regret in online convex optimization} \label{sec:gstar}

This section defines the $G^{\star}$ regret and compares it with the existing problem-dependent regret bounds. We also show that several existing {\oco} algorithms achieve $\mathcal{O} (
\sqrt{G^{\star}})$ regret, accompanied by a matching lower bound. The following assumptions will be frequently invoked throughout the paper.

\begin{enumerate}[leftmargin=25pt,itemsep=0pt,label=\textbf{A\arabic*:},ref=\rm{\textbf{A\arabic*}},start=1]
\item The losses $\ell_t : \mathbb{R}^n \rightarrow \mathbb{R}$ are $L$-smooth and convex \label{A1}
\begin{equation}
\ell_t (x) - \ell_t (y) - \langle \nabla \ell_t (y), x - y \rangle \geq \tfrac{1}{2 L} \| \nabla \ell_t (x) - \nabla \ell_t (y) \|^2 \quad\text{for all } x, y \in \Rbb^n. \label{eqn:Lsmooth}
\end{equation}
\item The closed convex set $\Xcal$ is bounded with $\diam (\mathcal{X}) \leq D$. \label{A2}
\end{enumerate}

\ref{A1} looks slightly different from the standard smoothness assumption $\ell_t (x) - \ell_t (y) - \langle \nabla \ell_t (y), x - y \rangle \leq \frac{L}{2}\|x-y\|^2$: an $L$-smooth convex function on $\Rbb^{n}$ is $\frac{1}{L}$-strongly convex in the dual space. It yields a tighter lower bound than that provided by convexity alone \cite{nesterov2018lectures}. This sharper lower bound is central to the derivation of the $G^\star$ regret.

\begin{rem}
The derivation of \ref{A1} from $L$-smoothness requires $\ell_t$ to be defined over the whole space. This assumption can be slightly relaxed to the case where $\Xcal \subsetneq \text{int} \dom\ell_t$. We discuss this extension in the appendix. 
\end{rem}

\subsection{$G^{\star}$ regret bound}

The definition of the $G^{\star}$ regret resembles that of the $L^{\star}$ regret,
replacing the optimality gap of the loss $\ell_t (x) - \inf \ell_t
$ by the non-stationarity $\| \nabla \ell_t (x) \|^2$:

\begin{definition}
  Given time horizon $T$ and a sequence of smooth convex losses $\{ \ell_t
  \}_{t = 1}^T$, the cumulative non-stationarity $G_T$ and the corresponding $G^{\star}$ regret are respectively defined as 
\begin{equation} \label{eqn:gstar}
\textstyle G_T(x) \assign \sum_{t = 1}^T \| \nabla
     \ell_t (x) \|^2 \quad  \text{and} \quad  G^{\star}_T \assign G_T(x^\star) =  \sum_{t = 1}^T \| \nabla
     \ell_t (x^\star) \|^2,
\end{equation}
where $x^\star \in \argmin_{x \in \Xcal} \sum_{t=1}^T \ell_t (x)$ is the minimizer of the cumulative losses.
\end{definition}

The $G^\star$ regret inherits the translation invariance of the gradient and does not require the losses to be non-negative or lower bounded. In particular, for online linear optimization with $\ell_t(x) = \langle g_t,x \rangle$, we have \[\textstyle G_T^\star \equiv G_T(x) = \sum_{t=1}^T \|\nabla \ell_t(x)\|^2= \sum_{t=1}^T \|g_t\|^2,\]and the $G^\star$ regret reduces to cumulative (sub)gradient norms, a quantity that many existing {\oco} algorithms adapt to. In contrast, the standard $L^\star$ regret is not well-defined in this setting.

\begin{rem}
Although the {\oco} problem is constrained, the definition of the $G^\star$ regret does not involve the constraint set. It is consistent with the way that the $L^\star$ regret is defined: in the literature, a lower bound on $\ell_t$ over $\Rbb^n$ (often $0$) is used in place of $\ell_t$, and the $L^\star$ regret often comes in the shifted form $\sum_{t=1}^T [\ell_t (x) - \inf \ell_t]\leq \sum_{t=1}^T \ell_t (x)$.
\end{rem}

\begin{rem}
It is feasible to define the $G^{\star}$ regret via the gradient-norm minimization problem
$\min_{x \in \mathcal{X}} G_T(x)$. However, the minimizer of $G_T$ need not coincide with
the point $x^{\star}$ that minimizes the cumulative losses $\sum_{t=1}^T \ell_t$, and the resulting regret bounds would not be directly comparable to $L_T^{\star}$. For ease of comparison, we define $G_T^{\star}$ as the $x^{\star}$ that minimizes cumulative losses. This choice suffices to
demonstrate the claimed tightness results.
\end{rem}

\paragraph{Connection with existing regret.}
The $G^\star$ regret measures the cumulative deviation from the unconstrained stationarity condition $\nabla \ell_t(x) = 0$. It lower bounds the $L^{\star}$ (assuming it exists) regret by \emph{self-boundedness} \cite{srebro2010smoothness}:
\begin{equation} \label{eqn:self-bound}
 \textstyle \| \nabla \ell_t (x) \|^2 \leq 2 L [\ell_t (x) - \inf 
\ell_t ].
\end{equation}

\Cref{prop:connection} relates the $G^\star$ regret to the existing problem-dependent regret bounds.
\begin{prop}\label{prop:connection}
  Under \ref{A1} and \ref{A2}, we always have $G^{\star}_T \leq 2 L (L^{\star}_T)$.
  Moreover, there exist loss sequences
  \begin{itemize}[leftmargin=15pt]
    \item  $\{ \ell_t^A \}_{t = 1}^T$ such that $G^{\star}_T \ll L_T^{\star}$;
    \item $\{ \ell_t^B \}_{t = 1}^T$ such that $V_T \ll L^{\star}_T ~(\text{or }G^{\star}_T)$;
    \item $\{ \ell_t^C \}_{t = 1}^T$ such that $V_T \gg L^{\star}_T ~(\text{or }G^{\star}_T)$.
  \end{itemize}
\end{prop}

\Cref{prop:connection} shows that small $L^\star$ regret implies small $G^\star$ regret, but the converse may not hold. Moreover, the $G^\star$ (or $L^\star$) regret and the gradient variation bound cannot imply each other. The remainder of this paper focuses on the $L^\star$ and $G^\star$ regret. Below are examples where the $G^\star$ regret can be substantially smaller than $L^\star$.

\begin{exple}[Logistic regression with cross-entropy loss] Consider logistic regression with cross-entropy loss $\ell_t (x) = \log (1 + 
  \exp (-y_t\langle a_t, x \rangle))$, where $\| a_t \| = 1$ is a data vector and $y_t \in \{1,-1\}$. Then
  $\nabla \ell_t (x) = \tfrac{-y_t a_t}{1 + \exp (y_t \langle a_t, x \rangle)}$ and $\|
  \nabla \ell_t (x) \|^2 \leq \ell_t (x)^2 = o (\ell_t (x))$ whenever $\ell_t
  (x) \rightarrow 0$. The observation generalizes to multi-class classification.
\end{exple}

\begin{exple}[Linear regression with $\ell_p$ loss]
	Consider online linear regression with loss $\ell_t(x) = \frac{1}{p}|r_t(x)|^p$ for $p \geq 2$, where the residual $r_t(x) = \langle a_t, x\rangle - b$. This loss is flat near its optimum, and the squared gradient norm vanishes faster than the function value: $\nabla \ell_t(x)  = a_t|r_t(x)|^{p-2} r_t(x)$ and $\tfrac{\|\nabla \ell_t(x)\|^2}{\ell_t(x)} = p\|a_t\|^2 |r_t(x)|^{p-2} = p^\frac{2p-2}{p}\|a_t\|^2 \ell_t(x)^\frac{p-2}{p}$. Then $\|\nabla \ell_t(x)\|^2 = o(\ell_t(x))$ whenever $r_t(x)\rightarrow 0$.
\end{exple}

\begin{exple}[Exponential loss]
	  Consider exponential loss $\ell_t (x) = \mathe^{-  \langle a_t, x \rangle}$, where $\|a_t\| = 1$ is a data vector. Then $\nabla \ell_t
  (x) = -\ell_t (x) a_t$ and $\| \nabla \ell_t (x) \|^2 = \ell_t (x)^2 = o
  (\ell_t (x)) $ whenever $\ell_t (x) \rightarrow 0$.
\end{exple}

\begin{exple}[Online regression with vanishing losses] \label{exp:loss}
Consider online regression problem with loss $\ell_t (x) = \frac{1}{2}(a_t x)^2, x \in [1,2]$. Suppose $a_t = t^{- p}$ for $p > 0$. Then
  $\nabla \ell_t (x) = a_t^2 x$ and $\| \nabla \ell_t (x) \|^2 = t^{- 4 p} x^2 = 2t^{-2
  p} \ell_t (x)$. Hence $\| \nabla \ell_t (x) \|^2 = o (\ell_t (x))$ as $t
  \rightarrow \infty$. In particular, when $p < \frac{1}{4}$, $L_T^\star = \frac{1}{2}\sum_{t=1}^Ta_t^2=\Theta(T^{1-2p})$ and $G_T^\star = \Theta(T^{1-4p})$.
\end{exple}

The above examples share the feature that the loss function becomes increasingly flat in a neighborhood of the optimum. For such loss functions, the $G^\star$ regret is typically much tighter than the $L^\star$ regret in the (near-)interpolation regime where $x^\star$ is nearly optimal for each of the losses.

\begin{rem} When the losses satisfy a lower curvature bound such as $\mu$-strong convexity, the quantities $L_T^\star$ and $G_T^\star$ can be
related via $2\mu L_T^\star \leq G_T^\star \leq 2L L_T^\star$. Our main focus is the regime where $\mu$ is close to zero, for which the $\Ocal(\tfrac{1}{\mu}\log T)$-type regret bounds become uninformative. For completeness, \Cref{sec:strongly-convex} establishes a logarithmic $G^\star$ regret bound when $\mu > 0$.
\end{rem}

\begin{rem}
The $L^\star$ regret can be obtained by directly assuming self-boundedness of the losses \eqref{eqn:self-bound}. Self-boundeness alone is weaker than $L$-smoothness and covers certain nonsmooth functions \cite{orabona2019modern}. On the other hand, it is unclear whether \eqref{eqn:Lsmooth} can be generalized to handle nonsmooth losses. We leave further investigations to future work.
\end{rem}

\subsection{Upper and lower bound for the $G^\star$ regret} \label{sec:gstar-ublb}

This section presents algorithms that achieve an $\mathcal{O} ( \sqrt{G^{\star}_T} )$ regret upper bound with a matching lower bound. The key idea is simple. Given any online algorithm that achieves $\Ocal(D \sqrt{\sum_{t=1}^T\|\nabla \ell_t(x^t)\|^2}) $ regret on the linearized losses (e.g. \cite{vaswani2020adaptive,streeter2010less}), applying \ref{A1} yields
\begin{align}
 \rho_T(x) ={}  \textstyle\sum_{t=1}^T \ell_t(x^t) - \ell_t(x)  \leq{} &  \textstyle\sum_{t=1}^T \langle \nabla \ell_t(x^t), x^t-x\rangle  - \frac{1}{2L}\sum_{t=1}^T\|\nabla \ell_t(x^t) - \nabla \ell_t(x)\|^2 \nonumber \\
 \leq{} & \textstyle \Ocal(D \sqrt{\sum_{t=1}^T\|\nabla \ell_t(x^t)\|^2}) - \frac{1}{2L}\sum_{t=1}^T\|\nabla \ell_t(x^t) - \nabla \ell_t(x)\|^2. \label{eqn:adp-convert}
\end{align}
\Cref{lem:sequence} below converts the  regret upper bound of type \eqref{eqn:adp-convert} into the desired $\Ocal(\sqrt{G_T(x)})$ regret guarantee by plugging in $a_t = \nabla \ell_t(x^t)$ and $b_t = \nabla \ell_t(x)$.
\begin{lem} \label{lem:sequence}
  Let $\{ a_t \}_{t = 1}^T$ and $\{ b_t \}_{t = 1}^T$ be two sequences of vectors
  and suppose $\alpha, \beta > 0$. Then
  \[ \textstyle \alpha \sqrt{\sum_{t = 1}^T \| a_t \|^2} - \beta \sum_{t = 1}^T \| a_t -
     b_t \|^2 \leq \alpha \sqrt{\sum_{t = 1}^T \| b_t \|^2} +
     \tfrac{\alpha^2}{4 \beta}.\]
\end{lem}

\begin{thm}
  \label{thm:master}Under \ref{A1}, any online algorithm that satisfies
  \[\textstyle \sum_{t = 1}^T \langle \nabla \ell_t (x^t), x^t - x \rangle \leq \delta (x)
  \sqrt{\sum_{t = 1}^T \| \nabla \ell_t (x^t) \|^2}\] achieves $\rho (x) \leq
  \delta (x) \sqrt{G_T (x)} + \frac{L}{2} \delta (x)^2$ regret for all $x \in \Xcal$.
\end{thm}

\Cref{thm:master} can convert existing results from \texttt{OCO} into their $G^\star$ version.

\paragraph{Lower bound.} Finally, since $G^\star_T$ is equivalent to cumulative squared gradient norm for linear losses, the $\Omega(\sqrt{T})$ lower bound for online linear optimization implies an $\Omega (
\sqrt{G^{\star}_T} )$ regret lower bound.

\begin{thm}
  \label{thm:lb}Let $T \geq 1$ and $M > 0$, and let $\Xcal$ be a closed convex set with $\diam (\Xcal) = D > 0$. For every possibly randomized online algorithm, there exists an oblivious sequence of linear losses $\ell_t (x) = \langle g_t, x \rangle$, with $\| g_t \| = M$, such that
  \[
    \textstyle \mathbb{E} [\rho_T (x^\star)] \geq \tfrac{D}{4} \sqrt{G_T^\star}.
  \]
  The expectation is over the internal randomization of the algorithm. For a deterministic algorithm, the inequality holds without expectation. The constructed instance satisfies \ref{A1} and \ref{A2}.
\end{thm}

%% file: sec_extension.tex
\section{Implications} \label{sec:extension}

This section showcases implications of the $G^\star$ regret to other {\oco} settings.

\subsection{Adaptive and unconstrained online learning}

\begin{coro}
  \label{thm:adanorm-gstar}Under \ref{A1} and \ref{A2}, {\adanorm} \cite{streeter2010less} \ifthenelse{\boolean{algolink}}{(\Cref{alg:adagrad-norm} in the appendix)}{(\textbf{Algorithm 2} in the appendix)}
  $x^{t + 1} = \Pi_{\mathcal{X}}
  [x^t - \eta_t \nabla \ell_t (x^t)]$ with $\eta_t =
   \frac{\alpha}{\sqrt{\sum_{i = 1}^t \| \nabla \ell_i (x^i) \|^2}}$ and $\alpha = \tfrac{\sqrt{2}D}{2}$ satisfies $\textstyle \rho_T (x)\leq \sqrt{2}  \sqrt{G_T(x)}D + L D^2$ for any $x \in \Xcal$.
\end{coro}

\begin{coro}
  \label{thm:adaftrl-gstar}Under \ref{A1} and \ref{A2}, suppose {\adaftrl} \cite{orabona2018scale} \ifthenelse{\boolean{algolink}}{(\Cref{alg:adaftrl} in the appendix)}{(\textbf{Algorithm 3} in the appendix)}
   is applied to the linearized losses $\hat{\ell}_t (x) \assign \langle \nabla \ell_t
  (x^t), x\rangle$. Further assume that {\adaftrl} uses a $\lambda$-strongly convex regularizer $r$ with $\lambda \geq \frac{1}{2D^2}$ and $\max_{x \in
  \mathcal{X}} r (x) + 1 =: R$. Then it satisfies $\textstyle \rho_T (x) \leq  \sqrt{3} R \sqrt{G_T(x)} D + 2 R^2 L D^2$ for any $x \in \Xcal$.
\end{coro}

\begin{coro}
  \label{coro:unconstrained-gstar}Under \ref{A1}, suppose unconstrained online learning algorithm \cite{cutkosky2024fully}
   is applied to the linearized losses $\hat{\ell}_t (x) \assign \langle \nabla \ell_t
  (x^t), x\rangle$. Then $\rho_T(x) \leq \Ocal(\|x\|\sqrt{G_T(x)} + \max_{t\in[T]} \|\nabla\ell_t(x^t)\|^2 + (L+1)\|x\|^2)$ for any $x \in \Rbb^n$.
\end{coro}

\subsection{Dynamic $G^\star$ regret} \label{sec:dynamic}
\input{sec_dynamic.tex}

\subsection{$G^\star$ regret under bandit feedback}  \label{sec:bandit}
\input{sec_bandit.tex}

%% file: sec_dynamic.tex
\Cref{sec:gstar} focuses on the static regret with respect to a single
competitor $x \in \mathcal{X}$. In this subsection, we consider dynamic regret,
which allows for a sequence of competitors at the cost of a path length term in the final regret bound. Given a comparator sequence $\{ \hat{x}_t \}_{t = 1}^T$ with path length $\textstyle \mathcal{P}_T (\{ \hat{x}_t \}_{t =
1}^T) \assign \sum_{t = 2}^T \| \hat{x}_{t + 1} - \hat{x}_t \|$,
the dynamic regret with respect to this sequence is defined as
\[ \textstyle \hat{\rho} (\{ \hat{x}_t \}_{t = 1}^T) \assign \sum_{t = 1}^T \ell_t (x^t)
   - \sum_{t = 1}^T \ell_t (\hat{x}_t) . \]
As in the
static regret setting, problem-dependent dynamic regret bounds have been established for both the $L^{\star}$ regret and the gradient variation bound
{\cite{zhao2024adaptivity,zhao2020dynamic}}. In particular, the
dynamic version of the $L^{\star}$ regret is given by $\hat{L}_T (\{ \hat{x}_t
\}_{t = 1}^T) \assign \sum_{t = 1}^T [\ell_t (\hat{x}_t) - \inf \ell_t]$. Online gradient descent with
constant learning rate $\eta$ achieves an $\mathcal{O} ( \tfrac{1
+\mathcal{P}_T}{\eta} + \eta \hat{L}_T )$ regret bound
{\cite{zhao2020dynamic}}. We now consider a dynamic analog of the $G^{\star}$ regret:
\[ \textstyle \hat{G}_T (\{ \hat{x}_t \}_{t = 1}^T) \assign \sum_{t = 1}^T \| \nabla
   \ell_t (\hat{x}_t) \|^2 . \]
As in the static regret case, self-boundedness implies $\hat{G}_T(\{ \hat{x}_t
\}_{t = 1}^T) \leq 2 L \hat{L}_T(\{ \hat{x}_t
\}_{t = 1}^T)$ given the same competitor sequence. \Cref{thm:dynamic-gstar-adaptive} below establishes a dynamic regret bound in
$\hat{G}_T$. 
%\begin{thm}% [Dynamic $G^\star$ regret]
%  \label{thm:dynamic-gstar}Under \ref{A1} and \ref{A2}, online
%  gradient descent with $\eta \in ( 0, \tfrac{1}{4L} ]$ satisfies $\hat{\rho} (\{ \hat{x}_t \}_{t = 1}^T) \leq \tfrac{D (D +2\mathcal{P}_T)}{2 \eta} + \eta
%     \hat{G}_T$ 
%  for any comparator sequence $\{ \hat{x}_t \}_{t = 1}^T,~\hat{x}_t \in
%  \mathcal{X}$. In particular, taking $\eta = \min\{\sqrt{\tfrac{{D (D
%  +2\mathcal{P}_T})}{{2 \hat{G}_T}}},\frac{1}{4L}\}$ gives $\hat{\rho} (\{ \hat{x}_t \}_{t =
%  1}^T) \leq \max\{ 2L D(D+2 \Pcal_T),\sqrt{{2 D (D +2\mathcal{P}_T) \hat{G}_T}}\}  =\mathcal{O} (
%  \sqrt{(1 + \Pcal_T + \hat{G}_T)(1 +\mathcal{P}_T) }  )$.
%\end{thm}
%Again, \Cref{thm:dynamic-gstar} is restrictive since the
%learning rate $\eta$ has to be chosen based on unknown quantities $\mathcal{P}_T$
%and $\hat{G}_T$.  This issue can be addressed by adopting the meta-expert algorithm framework in {\cite{zhao2024adaptivity}}. 
\begin{thm}
  \label{thm:dynamic-gstar-adaptive}Under \ref{A1} and \ref{A2}, there exists an online algorithm that achieves
  $\mathcal{O} ( \sqrt{(1 +\mathcal{P}_T + \hat{G}_T) (1 +\mathcal{P}_T)} \log \log T)$ dynamic regret without knowing $\hat{G}_T$ or $\Pcal_T$.
\end{thm}

The algorithm description and the proof of \Cref{thm:dynamic-gstar-adaptive} are left to
the appendix.

%% file: sec_bandit.tex
This subsection considers the setting in which access to the loss sequence $\{\ell_t\}_{t=1}^T$ is restricted to the two-point function-value feedback model \cite{flaxman2005online,duchi2015optimal,shamir2017optimal,he2025non}. We assume that the loss sequence is fixed independently of the learner's random directions. We use the two-point query model in which $x^t \in \Xcal$ is the decision at which regret is incurred, while the two function-value queries may lie outside $\Xcal$. These queries are well-defined because \ref{A1} assumes that every loss is defined on $\Rbb^n$. At each time step $t$, we draw a random direction $s_t$ uniformly from the unit sphere $\Sbb_n$ and query the loss at two points: $(y^t_{+}, y^t_{-}) = ( x^t + \mu s_t, x^t - \mu s_t)$, where $\mu > 0$ is called the smoothing parameter. After observing $\ell_t(y^t_{+})$ and
$\ell_t(y^t_{-})$, we construct the gradient estimator 
\begin{equation}\label{eq:gradient estimator}
      \textstyle g_t \assign \frac{n}{2\mu}\bigl[\ell_t(y^t_{+}) - \ell_t(y^t_{-})\bigr] s_t,
\end{equation}
and perform \textit{bandit gradient descent} (\bgd): $x^{t+1} = \Pi_{\Xcal}[x^t - \eta_t g_t]$. Let $\mathbb{B}_n \assign \{ u \in \Rbb^n : \| u \| \leq 1 \}$ denote the unit ball. The estimator $g_t$ is not an unbiased estimator of the true gradient $\nabla \ell_t(x^t)$; instead, it is an unbiased estimator of the gradient of $\tilde{\ell}_t$, the uniform-ball smoothing of $\ell_t$, at $x^t$:
\begin{equation}\label{eq:smoothing function}
\E_{s_t \sim \Sbb_n} [g_t | x^t] = \nabla \tilde{\ell}_t (x^t), \quad \text{where} \quad \tilde{\ell}_t (x) \assign \Ebb_{u_t \sim \tmop{Unif} (\mathbb{B}_n)} [\ell_t (x + \mu u_t)].
\end{equation}
This smoothed function $\tilde{\ell}_t$ naturally serves as a surrogate function when performing {\bgd}. Key properties of the smoothed function used in our analysis are summarized in the appendix. The following theorem establishes regret bounds for {\bgd} under an adaptive learning rate.

%\begin{thm}\label{thm:constant-bgd}
%    Under \ref{A1}, \ref{A2} and suppose $n \geq 8$, bandit gradient descent \ifthenelse{\boolean{algolink}}{(\Cref{alg:bgd} in the appendix)}{(\textbf{Algorithm 6} in the appendix)} $x^{t + 1} = \Pi_{\mathcal{X}} [x^t - \eta g_t]$ with constant learning rate $\eta \in ( 0, \tfrac{1}{4nL})$ satisfies $\Ebb [\rho_T(x)] \leq \textstyle\tfrac{D^2}{2\eta} + \tfrac{4n\eta}{1 - 4n\eta L}G_T(x) + \tfrac{4n\eta}{1 - 4n\eta L} T L^2\mu^2 + \tfrac{\eta T}{2}n^2 L^2 \mu^2 + \tfrac{TL\mu^2}{2}$ for any $x \in \mathcal{X}$. In particular, choosing $x = x^\star$, $\eta = \min \big\{
%  \tfrac{D}{4\sqrt{n G^{\star}_T}}, \tfrac{1}{8nL} \big\}$ and $\mu = \tfrac{D}{\sqrt{2nT}L}$ gives $ \E [\rho_T (x^\star)]\leq \max \{8n LD^2, 4D \sqrt{nG^{\star}_T} \} + \tfrac{L D^2}{2}$.
%\end{thm}

\begin{thm}
\label{thm:adaptive-bgd}
Suppose \ref{A1}, \ref{A2} hold and $n \geq 8$. Then bandit gradient descent \ifthenelse{\boolean{algolink}}{(\Cref{alg:adagrad-norm bgd} in the appendix)}{(\textbf{Algorithm 7} in the appendix)} $x^{t + 1} = \Pi_{\mathcal{X}} [x^t - \eta_t g_t]$ with {\adanorm} learning rate $\eta_t = \tfrac{\alpha}{\sqrt{\sum_{i=1}^t \|g_i\|^2}}$, $\alpha = \tfrac{\sqrt{2}D}{2}$ and $\mu = \tfrac{D}{n\sqrt{T}}$ satisfies $\E [\rho_T (x)] \leq \max\{16LD^2n, 4D\sqrt{n G_T (x)}\} + 5LD^2$ for any $x \in \mathcal{X}$, where the expectation is over the random directions $\{ s_t \}_{t = 1}^T$.
\end{thm}

%% file: sec_application.tex
\section{Applications and extensions} \label{sec:implication}

\subsection{Strongly convex and H\"{o}lder smooth losses} \label{sec:holder} \label{sec:strongly-convex}

This subsection considers the generalization of \ref{A1} and the consequences of additional curvature assumptions.

\paragraph{Strong convexity.}
\input{sec_strongly_convex.tex}

\paragraph{H\"{o}lder smoothness.}
\input{sec_holder.tex}

\subsection{Stochastic optimization under interpolation and heavy-tail noise}
\input{sec_sgd.tex}

\subsection{Constraint-aware $G^{\star}$ regret via smooth surrogate loss}
\input{sec_smoothing.tex}

%% file: sec_strongly_convex.tex
We first show that $G^\star$ regret generalizes to strongly convex losses.
\begin{enumerate}[leftmargin=25pt,itemsep=0pt,label=\textbf{A\arabic*:},ref=\rm{\textbf{A\arabic*}},start=4]
\item The losses $\ell_t : \Rbb^n \rightarrow \Rbb$ are $\mu$-strongly convex for some $0 < \mu \leq L$: \label{A4}
\begin{equation} \label{eqn:strong-convex}
  \textstyle \ell_t (x) - \ell_t (y) - \langle \nabla \ell_t (y), x - y \rangle \geq \frac{\mu}{2} \| x - y \|^2 \quad \text{for all } x, y \in \Rbb^n.
\end{equation}
\end{enumerate}
Under \ref{A1} and \ref{A4}, $2 \mu L_T^{\star} \leq G_T^{\star} \leq 2 L L_T^{\star}$. Substituting $L_T^{\star} \leq G_T^{\star} / (2 \mu)$ into an existing $\log (L L_T^{\star})$ bound places the condition number $L / \mu$ inside the comparator-dependent logarithm. To remove this dependence, the following logarithmic counterpart of \Cref{lem:sequence} gives a direct bound in terms of $G_T^{\star}$.

\begin{lem} \label{lem:log-sequence}
  Let $\{ a_t \}_{t = 1}^T$ and $\{ b_t \}_{t = 1}^T$ be two sequences of vectors, and suppose that $\alpha, \beta, \gamma > 0$. Then
  \[
    \textstyle \alpha \log (1 + \gamma \sum_{t = 1}^T \| a_t \|^2) - \beta \sum_{t = 1}^T \| a_t - b_t \|^2 \leq \alpha \max \{ \log (1 + 2 \gamma \sum_{t = 1}^T \| b_t \|^2), \log (\frac{2 \alpha \gamma}{\beta}) \}.
  \]
\end{lem}

The next result uses a slightly more conservative learning rate than the one in the original analysis of smooth and strongly convex online gradient descent (\texttt{S2OGD}) \cite{wang2020adapting}. This retains a gradient-difference penalty from \ref{A1} for the logarithmic conversion.

\begin{thm}
  \label{thm:strongly-convex-gstar}Suppose \ref{A1}, \ref{A2}, and \ref{A4} hold, and the parameters $L$ and $\mu$ are known. Let $M > 0$ be a known constant such that $\| \nabla \ell_t (u) \| \leq M$ for all $t \in [T]$ and $u \in \Xcal$. Starting from any $x^1 \in \Xcal$, consider \texttt{S2OGD} \ifthenelse{\boolean{algolink}}{(\Cref{alg:s2ogd} in the appendix)}{(\textbf{Algorithm 4} in the appendix)} with
  \[
    \textstyle x^{t + 1} = \Pi_{\Xcal} [x^t - \eta_t \nabla \ell_t (x^t)] \quad \text{and} \quad \eta_t = \frac{2 M^2 / \mu}{8 L M^2 / \mu + \sum_{i = 1}^t \| \nabla \ell_i (x^i) \|^2}.
  \]
  Then $\textstyle \rho_T (x) \leq 2 L D^2 + \frac{M^2}{\mu} \log (1 + \frac{\mu G_T (x)}{4 L M^2})$ for any $x \in \Xcal$.
\end{thm}

Since $G_T^{\star} \leq M^2 T$, \Cref{thm:strongly-convex-gstar} recovers the $\mathcal{O} (L D^2 + \frac{M^2}{\mu} \log T)$ regret. Moreover, using $G_T^{\star} \leq 2 L L_T^{\star}$ gives
\[
  \textstyle \rho_T (x^{\star}) \leq 2 L D^2 + \frac{M^2}{\mu} \log (1 + \frac{\mu L_T^{\star}}{2 M^2}),
\]
which recovers the existing $\mathcal{O} (\log L_T^{\star})$ dependence up to constants. Thus, the comparator-dependent quantity is no worse than in the $L^{\star}$ bound. The retuned learning rate incurs the comparator-independent term $2 L D^2$, and the possible improvement from $G_T^{\star}$ occurs only inside a logarithm.

%% file: sec_holder.tex
We further extend the $G^{\star}$ regret analysis to convex H\"{o}lder-smooth losses. We replace \ref{A1} with the following weaker condition.
\begin{enumerate}[leftmargin=25pt,itemsep=0pt,label=\textbf{A\arabic*:},ref=\rm{\textbf{A\arabic*}},start=3]
\item Fix $\nu \in (0,1]$. The losses $\ell_t : \Rbb^n \rightarrow \Rbb$ are convex and ($L$, $\nu$)-H\"{o}lder smooth in the following sense: \label{A3}
\begin{equation} \label{eqn:holder-smooth}
  \textstyle \ell_t (x) - \ell_t (y) - \langle \nabla \ell_t (y), x - y \rangle \geq \frac{\nu}{1 + \nu} L^{-1 / \nu} \| \nabla \ell_t (x) - \nabla \ell_t (y) \|^{\frac{1 + \nu}{\nu}} \quad \text{for all } x, y \in \Rbb^n.
\end{equation}
\end{enumerate}
By convexity, the standard H\"{o}lder-smoothness condition $\| \nabla \ell_t (x) - \nabla \ell_t (y) \| \leq L \| x - y \|^\nu$ implies the inequality in \ref{A3}; see Lemma 4.2 of \cite{he2025polyak}. When $\nu = 1$, \ref{A3} reduces to \ref{A1}. Then we define the corresponding $G^\star$ quantities.

\begin{definition}
  Given time horizon $T$ and a sequence of ($L$, $\nu$) H\"{o}lder-smooth losses, the H\"{o}lder cumulative non-stationarity $G_{\nu,T}$ and the corresponding $G_{\nu,T}^{\star}$ regret are respectively defined as
  \begin{equation} \label{eqn:holder-gstar}
    \textstyle G_{\nu,T} (x) \assign \sum_{t = 1}^T \| \nabla \ell_t (x) \|^{\frac{1 + \nu}{\nu}} \quad \text{and} \quad G_{\nu,T}^{\star} \assign G_{\nu,T} (x^{\star}),
  \end{equation}
  where $x^{\star} \in \argmin_{x \in \Xcal} \sum_{t = 1}^T \ell_t (x)$.
\end{definition}

The self-boundedness property under \ref{A3} gives $\textstyle \| \nabla \ell_t (x) \|^{\frac{1 + \nu}{\nu}} \leq \frac{1 + \nu}{\nu} L^{1 / \nu} [\ell_t (x) - \inf \ell_t]$, and consequently, $G_{\nu,T}^{\star} \leq \frac{1 + \nu}{\nu} L^{1 / \nu} L_T^{\star}$ whenever $L_T^{\star}$ is well-defined. The following lemma is the H\"{o}lder counterpart of \Cref{lem:sequence}.

\begin{lem} \label{lem:holder-gstar-conversion}
  Let $\{ a_t \}_{t = 1}^T$ and $\{ b_t \}_{t = 1}^T$ be two sequences of vectors, and suppose $\alpha, \beta > 0, \nu \in (0,1]$. Then
  \[
    \textstyle \alpha \sqrt{\sum_{t = 1}^T \| a_t \|^2} - \beta \sum_{t = 1}^T \| a_t - b_t \|^{\frac{1 + \nu}{\nu}} \leq \alpha \sqrt{\sum_{t = 1}^T \| b_t \|^2} + \frac{\nu^\nu}{(1 + \nu)^{1 + \nu}} \frac{\alpha^{1 + \nu}}{\beta^\nu} T^{\frac{1 - \nu}{2}}.
  \]
\end{lem}

\begin{thm}
  \label{thm:holder-gstar}Under \ref{A3}, any online algorithm that satisfies $\textstyle \sum_{t = 1}^T \langle \nabla \ell_t (x^t), x^t - x \rangle \leq \delta (x) \sqrt{\sum_{t = 1}^T \| \nabla \ell_t (x^t) \|^2}$
  achieves the regret bound $\textstyle \rho_T (x) \leq \delta (x) \sqrt{G_T (x)} + \frac{L}{1 + \nu} \delta (x)^{1 + \nu} T^{\frac{1 - \nu}{2}}$
  for all $x \in \Xcal$.
\end{thm}

\begin{coro}
  \label{coro:holder-adanorm}Under \ref{A2} and \ref{A3}, {\adanorm} \cite{streeter2010less} \ifthenelse{\boolean{algolink}}{(\Cref{alg:adagrad-norm} in the appendix)}{(\textbf{Algorithm 2} in the appendix)} with $\alpha = \tfrac{\sqrt{2}D}{2}$ satisfies
$\textstyle \rho_T (x) \leq \sqrt{2}D \sqrt{G_T (x)} + \frac{L}{1 + \nu} (\sqrt{2}D)^{1 + \nu} T^{\frac{1 - \nu}{2}}$
  for any $x \in \Xcal$.
\end{coro}

When $\nu = 1$, \Cref{thm:holder-gstar} recovers the standard $L$-smooth $G_T (x)$ bound. When $\nu < 1$, the comparator-gradient term remains $\sqrt{G_T (x)}$, while the constant smoothness remainder is replaced by $\frac{L}{1 + \nu} \delta (x)^{1 + \nu} T^{\frac{1 - \nu}{2}}$. H\"{o}lder's inequality provides a bound directly in terms of $G_{\nu,T} (x)$.

\begin{coro} \label{coro:holder-gnu-upper}
  Under the same conditions as \Cref{thm:holder-gstar}, $\textstyle \rho_T (x) \leq \delta (x) T^{\frac{1 - \nu}{2 (1 + \nu)}} G_{\nu,T} (x)^{\frac{\nu}{1 + \nu}} + \frac{L}{1 + \nu} \delta (x)^{1 + \nu} T^{\frac{1 - \nu}{2}}$ for any $x \in \Xcal$.
\end{coro}

%% file: sec_sgd.tex
Consider the
following optimization problem 
\begin{equation}
  \min_{x \in \mathcal{X}} ~ f (x) \assign \mathbb{E}_{\xi \sim \Xi} [f (x,
  \xi)], \label{eqn:stocopt}
\end{equation}
where we assume that $f (x, \xi)$ is convex and $L$-smooth for all $\xi \sim
\Xi$. The convergence analysis of stochastic gradient descent relies on
assumptions on the stochastic noise, and the following two quantities, known
as {\tmem{function noise}} and {\tmem{gradient noise}} respectively
{\cite{garrigos2023handbook,gower2019sgd}}, are widely used in the literature.
\[ \textstyle \sigma_f^{\star} \assign \inf_x \mathbb{E} [f (x, \xi) - \inf_x f (x, \xi)]
   \quad \text{and} \quad \sigma_g^{\star} \assign \sqrt{\mathbb{E} [\| \nabla
   f (x^{\star}, \xi) \|^2]}, \]
where $x^{\star} \in \argmin_x f (x)$. We also make the following simplifying assumption.

\begin{enumerate}[leftmargin=25pt,itemsep=2pt,label=\textbf{A\arabic*:},ref=\rm{\textbf{A\arabic*}},start=5]
\item There is some unconstrained minimizer $x^\star \in \argmin_x f(x)$ such that $x^\star \in \Xcal$. \label{A5}
\end{enumerate}

Under \ref{A5}, we have $\sigma_f^\star = \min_{x\in \Xcal} \mathbb{E} [f (x, \xi) - \inf_x f (x, \xi)]$, and it is easy to verify  that $\sigma_f^{\star}$ and $\sigma_g^{\star}$ mirror the $L^{\star}$ and
$G^{\star}$ regret, respectively. When $\sigma_f^{\star}$ or $\sigma_g^{\star}$ is close to
$0$, the regime is known as (near-){\tmem{interpolation}}
{\cite{mishkin2020interpolation}}. Interpolation is a central setting in optimization for machine learning and has recently attracted significant attention
{\cite{srebro2010smoothness,ma2018power,vaswani2019painless,attia2025fast}}. Since online algorithms with sublinear regret can be converted into an optimization algorithm via online-to-batch conversion \cite{orabona2019modern}, our $G^{\star}$ regret bounds from \Cref{sec:gstar-ublb} imply convergence guarantees
for stochastic gradient methods and stochastic dual averaging (i.e., follow-the-regularized-leader for linearized losses).
\begin{coro} \label{coro:stochastic-opt}
  Suppose $f (x, \xi)$ is $L$-smooth and convex for all $\xi$ and that
  \ref{A2}, \ref{A5} hold. Let $\ell_t (x) = f (x, \xi^t)$ and $\{ \xi^t \}_{t =
  1}^T$ be i.i.d. samples drawn from $\Xi$. Then
  \begin{itemize}[leftmargin=12pt,itemsep=0pt,topsep=3pt]
    \item if {\ogd} is applied to
    $\{\ell_t\}_{t=1}^T$, then for any $\eta \in ( 0, \tfrac{1}{L}
    )$, we have
    \[ \mathbb{E} [f (\bar{x}^T) - f (x^{\star})] \leq \tfrac{1}{2 \eta T} \|
       x^1 - x^{\star} \|^2 + \tfrac{\eta}{2(1 - \eta L)} (\sigma_g^{\star})^2; \]
       \vspace{-15pt}
    \item if {\adanorm} from \Cref{thm:adanorm-gstar} is applied to $\{\ell_t\}_{t=1}^T$, then $\mathbb{E} [f (\bar{x}^T) - f (x^{\star})] \leq \tfrac{L D^2}{T} +
       \tfrac{\sqrt{2} D}{\sqrt{T}} \sigma_g^{\star}$.
    \item if {\adaftrl} from \Cref{thm:adaftrl-gstar} is applied to $\{\ell_t\}_{t=1}^T$, then $\mathbb{E} [f (\bar{x}^T) - f (x^{\star})] \leq \tfrac{2R^2 L D^2}{T} +
       \tfrac{\sqrt{3}R D}{\sqrt{T}} \sigma_g^{\star}$.
  \end{itemize}
where $\bar{x}^T \assign \tfrac{1}{T} \sum_{t = 1}^T x^t$ is the average
  iterate and $R$ is defined in \Cref{thm:adaftrl-gstar}.
\end{coro}

\begin{rem}
\Cref{coro:stochastic-opt} improves the existing results \cite{srebro2010smoothness} by replacing $\sigma_f^\star$ by $\sigma_g^\star$. Compared with the recent last-iterate convergence results of stochastic gradient descent \cite{attia2025fast,garrigos2025last}, our result enables adaptivity to smoothness for the average iterate (\Cref{tab:stochastic}).
\end{rem}
\vspace{-5pt}
\begin{table}[h]
\centering
\caption{Implications of the $G^\star$ regret for stochastic smooth convex optimization \label{tab:stochastic}}\vspace{-10pt}
\resizebox{0.9\textwidth}{!}{
  \begin{tabular}{ccccc}
    \toprule
    Reference & Setting & Algorithm & Adapt to $\sigma^{\star}$ & Rate\\
    \midrule
    {\cite{srebro2010smoothness}} & Average iterate,
    $\mathcal{X}=\mathbb{R}^n$  & \texttt{SGD} & No & $\mathcal{O} ( \tfrac{1}{T}
    + \tfrac{1}{\sqrt{T}} \sigma_f^{\star})$\\
    {\cite{attia2025fast,garrigos2025last}} & Last iterate,
    $\mathcal{X}=\mathbb{R}^n$ & \texttt{SGD} & No & $\mathcal{O} \big( \tfrac{\log
    T}{T} + \sqrt{\tfrac{\log T}{T}} \sigma_g^{\star} \big)$\\
    {\cite{vaswani2020adaptive}} & Average iterate, $\tmop{diam} (\mathcal{X})
    \leq D$ & {\adagrad} & Yes & $\mathcal{O} ( \tfrac{1}{T} +
    \tfrac{1}{\sqrt{T}} \sigma_f^{\star} )$\\
    This paper & Average iterate, $\diam (\mathcal{X}) \leq D$ &
    {\adanorm}, {\adaftrl} & Yes & $\mathcal{O} ( \tfrac{1}{T} +
    \tfrac{1}{\sqrt{T}} \sigma_g^{\star} )$\\
    \bottomrule
  \end{tabular}
}
\end{table}
\vspace{-5pt}
\paragraph{Improved analysis of {\adanorm}.} As a byproduct, we are also able to slightly refine the existing convergence analysis of {\adanorm} for unconstrained smooth convex optimization \cite{liu2022convergence}.

\begin{thm} \label{thm:adanorm-opt}
Suppose {\adanorm} is applied to \eqref{eqn:stocopt} with $\Xcal = \Rbb^n$ and that there exists $x^\star$ such that $\nabla f(x^\star, \xi) = 0$ surely. Further assume that each $f(x, \xi)$ is $L$-smooth.  Then $\Ebb[f(\bar{x}^{T}) - f(x^\star)] \leq \tfrac{L}{2 T} (
  \tfrac{\| x^1 - x^{\star} \|^2}{2 \alpha} + \alpha \log (
  \tfrac{\sqrt{\mathe}\alpha L}{\| \nabla f (x^1, \xi^1) \|} ) +
  \tfrac{1}{2 L} \| \nabla f (x^1, \xi^1) \| )^2$, where $\bar{x}^T \assign \tfrac{1}{T} \sum_{t = 1}^T x^t$ is the average iterate.
\end{thm}

\paragraph{Heavy-tailed noise.}
The last two guarantees in \Cref{coro:stochastic-opt} extend to heavy-tailed stochastic gradients with possibly infinite variance. For any $p \in (1,2]$, define $\sigma_{g,p}^{\star} \assign (\mathbb{E}_{\xi} [\| \nabla f (x^{\star}, \xi) \|^p])^{1 / p}$ and suppose that $\sigma_{g,p}^{\star} < \infty$. Under \ref{A5}, $\nabla f (x^{\star}) = 0$, so this is a bounded $p$-th moment of the stochastic-gradient noise at the optimum. Setting $z_t \assign \| \nabla f (x^{\star}, \xi^t) \|$, the inequality $\| z \| \leq \| z \|_p$ for $p \in (1,2]$, followed by Jensen's inequality, gives
\[
  \textstyle \mathbb{E} [\sqrt{G_T (x^{\star})}] \leq \mathbb{E} [(\sum_{t = 1}^T \| \nabla f (x^{\star}, \xi^t) \|^p)^{1 / p}] \leq (\sum_{t = 1}^T \mathbb{E} [\| \nabla f (x^{\star}, \xi^t) \|^p])^{1 / p} = T^{1 / p} \sigma_{g,p}^{\star}.
\]
Consequently, the last two bounds in \Cref{coro:stochastic-opt} remain valid after replacing $\sigma_g^{\star} / \sqrt{T}$ by $\sigma_{g,p}^{\star} / T^{1 - 1 / p}$. Thus, our $G^{\star}$ analysis requires a bounded noise moment only at $x^{\star}$, rather than a uniform moment bound over $\Xcal$ or along the iterates. This is the same $\ell_p$-to-$\ell_2$ comparison used to analyze {\adanorm} under heavy-tailed noise in Section 3.3 of \cite{liu2026online}. The fixed-step {\ogd} bound does not share this extension for $p < 2$, since it depends on the sum of the second moments.

%% file: sec_smoothing.tex
Although this paper focuses on constrained {\oco}, the
definition of $L^{\star}_T$ measures suboptimality over
$\mathbb{R}^n$ and does not take into account the constraint set
$\mathcal{X}$. A more desirable measure is
\[ \textstyle L^{\mathcal{X}}_T (x) \assign \sum_{t = 1}^T \ell_t (x) - \min_{u \in
   \mathcal{X}} \ell_t (u), \]
which measures suboptimality relative to the constrained optimum rather than the unconstrained optimum $\sum_{t = 1}^T \ell_t (x) - \inf \ell_t$.
% (or $\sum_{t = 1}^T \ell_t (x)$ for non-negative losses). 
Similarly for $G^{\star}$, the gradient
norm $\| \nabla \ell_t (x) \|$ is oblivious of the constraint set
$\mathcal{X}$.
Instead, a more appropriate measure should involve
$\dist (- \nabla \ell_t (x), \mathcal{N}_{\mathcal{X}} (x))$, where $\mathcal{N}_{\mathcal{X}} (x) \assign
\{ g : \langle g, x - y \rangle \geq 0, \text{ for all } y \in
\mathcal{X} \}$ denotes the normal cone of $\mathcal{X}$ at $x$. \\

The constrained cumulative suboptimality $L^{\mathcal{X}}_T$ can provide sharper problem-dependent constants. For example, suppose the unconstrained optimum is not in $\mathcal X$ but the constrained optimum
$x^{\star} \in \tmop{bd} \mathcal{X}$ is also the constrained optimum for most losses individually: $ \ell_t(x^\star) = \min_{x \in \Xcal} \ell_t(x)$. In this case, $L^{\mathcal{X}}_T (x^{\star}) = \mathcal{O} (1)$, whereas $L^\star_T$ can still be $\Omega (T)$. In the rest of this section, we demonstrate the possibility of generalizing
the existing small loss (gradient) bound to adapt to the constraint set
$\mathcal{X}$.

\paragraph{Hardness.}As discussed, the naive goal is to design an online algorithm that directly adapts to $L^{\mathcal{X}}_T$ (or $G^{\mathcal{X}}_T$) by replacing
the traditional $\mathcal{O} ( L D^2 + D \sqrt{L L^{\star}_T} )$ by
$\mathcal{O} ( L D^2 + D \sqrt{L L^{\mathcal{X}}_T (x^{\star})} )$.
However, \Cref{prop:hardness} shows that this goal is generally not
achievable.

\begin{prop}
  \label{prop:hardness}For any $\tau \in [ 0, \tfrac{1}{6} )$,
  there exists a loss sequence $\{ \ell_t \}_{t = 1}^T$ satisfying
  \ref{A1}, \ref{A2} and that $\inf \ell_t = 0$ such that for any
  online algorithm and $x \in \mathcal{X}$, \ $\frac{\rho_T (x)}{L D^2 + D
  \sqrt{L L^{\mathcal{X}}_T (x)}} = \Omega (T^{\tau})$.
\end{prop}

Despite the hardness result, whenever it is possible to redefine the losses,
we show a strategy that enables dependence on $L^{\mathcal{X}}_T$ and
$G^{\mathcal{X}}_T$ by replacing the losses by their smooth surrogates.

\paragraph{Solution: smooth indicator of constraints.} Given a lower-semicontinuous convex function
$\ell : \mathcal{X} \rightarrow \mathbb{R}$, its associated Moreau envelope with
parameter $\gamma > 0$, denoted by $\ell^{1 / \gamma} (x)$, is given by
\[ \textstyle \ell^{1 / \gamma} (x) \assign \min_{y \in \mathcal{X}}~ \{ \ell (y) +
   \tfrac{\gamma}{2} \| y - x \|^2 \} . \]
Regardless of the domain or smoothness of $\ell$, the function $\ell^{1 / \gamma}$ is always
$\gamma$-smooth and convex, with gradient given by $\nabla \ell^{1 / \gamma} (x)
= \gamma (x - \tmop{prox}_{\ell / \gamma} (x))$, where $\tmop{prox}_{\ell / \gamma}
(x) = \argmin_x \{ \ell (y) + \tfrac{\gamma}{2} \| y - x \|^2 \}$ is
the proximal point. In particular, $\ell^{1 / \gamma}$ and $\ell$ share the same
minimizers and stationarity points, and $\nabla \ell^{1 / \gamma} (x)$ approximates the quantity $\dist (- \nabla \ell (x),
\mathcal{N}_{\mathcal{X}} (x))$ \cite{deng2025smooth}. Hence, we similarly define
$G^{\mathcal{X}}_T (x) \assign \sum_{t = 1}^T \| \nabla \ell_t^{1 / \gamma} (x)
\|^2$. In view of the properties of the smooth approximation, instead of
solving the original {\oco} problem, we consider the {\oco} problem defined on the smooth surrogates:
\[ \textstyle \rho_T^{1 / \gamma} (x) \assign \sum_{t = 1}^T \ell^{1 / \gamma}_t (x^t) -
   \ell^{1 / \gamma}_t (x), \quad x^t \in \mathcal{X}, t \in[T]. \]
Since each loss $\ell^{1 / \gamma}_t$ is convex and $\gamma$-smooth, the results from the previous sections directly apply.
\begin{prop}\label{prop:smooth}
  We have $G^{\mathcal{X}}_T (x) \leq 2 \gamma
  L^{\mathcal{X}}_T (x)$, and the results from previous sections hold for $\rho^{1/\gamma}_T$. In particular, there exist online algorithms satisfying $\rho_T^{1 / \gamma} (x) = \mathcal{O} (\gamma D^2+
  \sqrt{G^{\mathcal{X}}_T (x)} ) = \mathcal{O} (\gamma D^2+
  \sqrt{\gamma L^{\mathcal{X}}_T (x)} )$ for any $x \in \mathcal{X}$.
\end{prop}
We make two remarks on running online learning algorithms on the surrogate loss
sequence $\{ \ell^{1 / \gamma}_t \}_{t = 1}^T$. First and importantly, the regret guarantee in \Cref{prop:smooth} is with respect to $\rho_T^{1 / \gamma}$ instead of
$\rho_T$, and the gap between $\ell^{1 / \gamma} (x^t)$ and $\ell (x^t)$ can be up to
$\mathcal{O} (\gamma^{- 1})$ \cite{beck2012smoothing, deng2025smooth}. In particular, using Proposition 4.2 of \cite{deng2025smooth}, we have
\begin{equation} \label{eqn:envelop-approx}
	\ell^{1/\gamma} (x) + \tfrac{1}{2\gamma} \|\nabla \ell^{1/\gamma}(x)\|^2\leq \ell(x) \leq \ell^{1/\gamma}(x) + \tfrac{1}{\gamma}\|\nabla \ell(x)\|^2,
\end{equation}
and combining \eqref{eqn:envelop-approx} with \Cref{prop:smooth} yields a bound on the regret:
\begin{equation} \label{eqn:smooth-regret}
	 \textstyle \rho_T(x) \leq \rho^{1/\gamma}_T(x) + \frac{1}{\gamma}\sum_{t=1}^T \|\nabla\ell_t(x^t) \|^2 \leq \Ocal\big(\gamma D^2 + \frac{1}{\gamma}\sum_{t=1}^T \|\nabla\ell_t(x^t) \|^2 + \sqrt{G_T^\Xcal(x)}\big).
\end{equation}
Assuming $\|\nabla\ell(x^t)\| \leq M$, the first two terms $\gamma D^2 + \frac{1}{\gamma} T M^2$ require
$\gamma =\Omega (\sqrt{T}
)$ to guarantee $\rho_T (x) \leq \mathcal{ O} ( \sqrt{T} )$, and there is no regret improvement in the worst-case. In the appendix, we show hardness of obtaining an $\Ocal(L D^2 + D
  \sqrt{G^{\mathcal{X}}_T (x)})$ on $\rho_T(x)$ using an argument similar to \Cref{prop:hardness}.
While this argument yields an improved bound for the smoothed loss, it does not always yield an improved bound for the original loss.
%Nevertheless, whenever it is feasible to directly redefine the loss, 
%the smoothed loss adapts to the constrained small loss bound $L^{\mathcal{X}}_T (x)$. 

\begin{rem}
The $G^\star$ regret can be strictly tighter than the $L^\star$ regret. To see this,
notice that for $\gamma = \Theta(\sqrt{T})$,
the $\mathcal{O} (\gamma D^2+
  \sqrt{\gamma L^{\mathcal{X}}_T (x)} )$ upper bound  yields a $\sqrt{\gamma L^{\mathcal{X}}_T (x)}$ term that can be as large as $\Theta (T^{3/4})$ in the worst case.
\end{rem}

Second, running online gradient descent on $\ell^{1 / \gamma}_t$ requires computing $\nabla \ell_t^{1 / \gamma}$. This quantity can be evaluated using the proximal point iteration, which is often efficient \cite{shtoff2024efficient}. 
Note that running an online algorithm on smooth surrogates $\{ \ell^{1 / \gamma}_t \}_{t = 1}^T$ differs from the existing implicit online learning
(online proximal point) approach \cite{kulis2010implicit}, since the proximal point is applied not as an online algorithm itself, but instead as a subroutine to find the envelope gradient $\nabla \ell_t^{1 / \gamma} $. This smoothing strategy can be applied jointly with any existing adaptive online algorithm, such as {\adagrad}.

%% file: sec_exp.tex
\section{Illustrative experiments} \label{sec:exp}

In this section, we provide toy numerical experiments to demonstrate the tightness
of the $G^{\star}$ regret bound.

\subsection{Experiment setup}

We consider two kinds of loss functions from \Cref{sec:gstar}.
\begin{itemize}[leftmargin=10pt]
  \item regression with $\ell_p$ loss $\ell_t^{p} (x) \assign
  \tfrac{1}{p} |\langle a_t, x \rangle - b_t|^p$, and
  \item binary classification with cross-entropy loss $\ell_t^{\mathrm{ce}} (x)
  \assign \log (1 + \exp (y_t \langle a_t, x \rangle))$ 
\end{itemize}
\paragraph{Data preparation.} We fix $n = 2$ for all the experiments.
 For $\ell_p$-regression, we let $p = 4$ and generate $a_t \sim \mathcal{N}
(0, \frac{1}{10}I_n)$ from standard normal distribution and generate $\bar{x} \sim
\mathcal{N} (0,\frac{1}{10}I_n)$. Then we let $b_t = \langle a_t, \bar{x} \rangle +
\sigma \varepsilon_t$ with $\varepsilon_t \sim \mathcal{N} (0, 1)$ and
$\sigma \in [0.1, 1]$ parametrizes the noise level. As $\sigma \rightarrow 0$,
$\ell_t^{p} (\bar{x}) \rightarrow 0$ for all $t \in [T]$.
For binary classification, we generate i.i.d. $a_t \sim \mathcal{U} [0, 1]^n$
from uniform distribution and $y_t \sim \tmop{Bernoulli} (\delta), \delta \in
[0.9, 1]$. As $\delta \rightarrow 1$, it is more likely to find $\bar{x}$ such
that $y_t \langle a_t, \bar{x} \rangle \rightarrow - \infty$ and
$\ell_t^{\mathrm{ce}} (x) \rightarrow 0$ for most of $t \in [T]$.
\paragraph{Algorithm configuration.}
After generating $\{ \ell_t \}_{t = 1}^T$, we compute $x^{\star} \in \arg
\min_{x \in \{ x : \| x \| \leq 10^{-2} \}}$ and obtain $L_T^{\star},
G_T^{\star}$ using \texttt{CVX} \cite{grant2008cvx}. Finally, we run {\adanorm} with parameter $\alpha = \frac{\sqrt{2}D}{2}$ and obtain $\rho_T (x^{\star})$. The range of time horizon is taken to be $T \in [10^2, 10^4]$.

\subsection{Regret comparison}

\begin{figure}[h]
\centering
\includegraphics[height=0.175\textwidth]{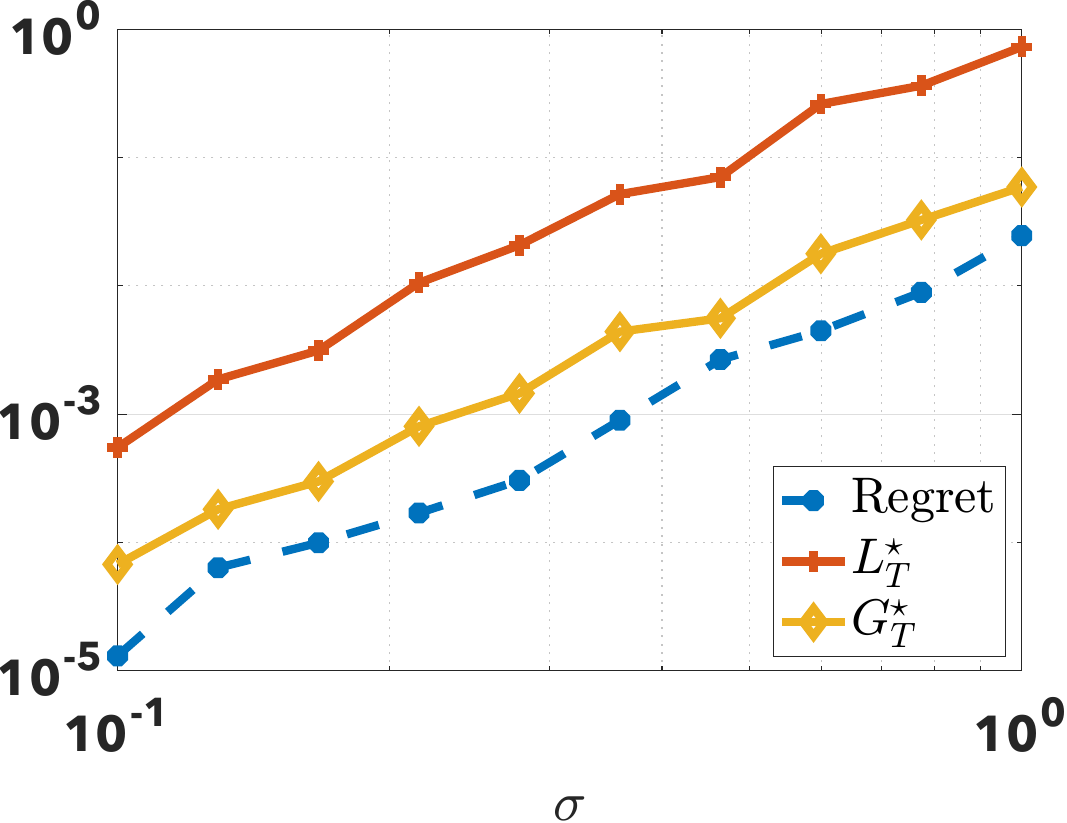}
\includegraphics[height=0.175\textwidth]{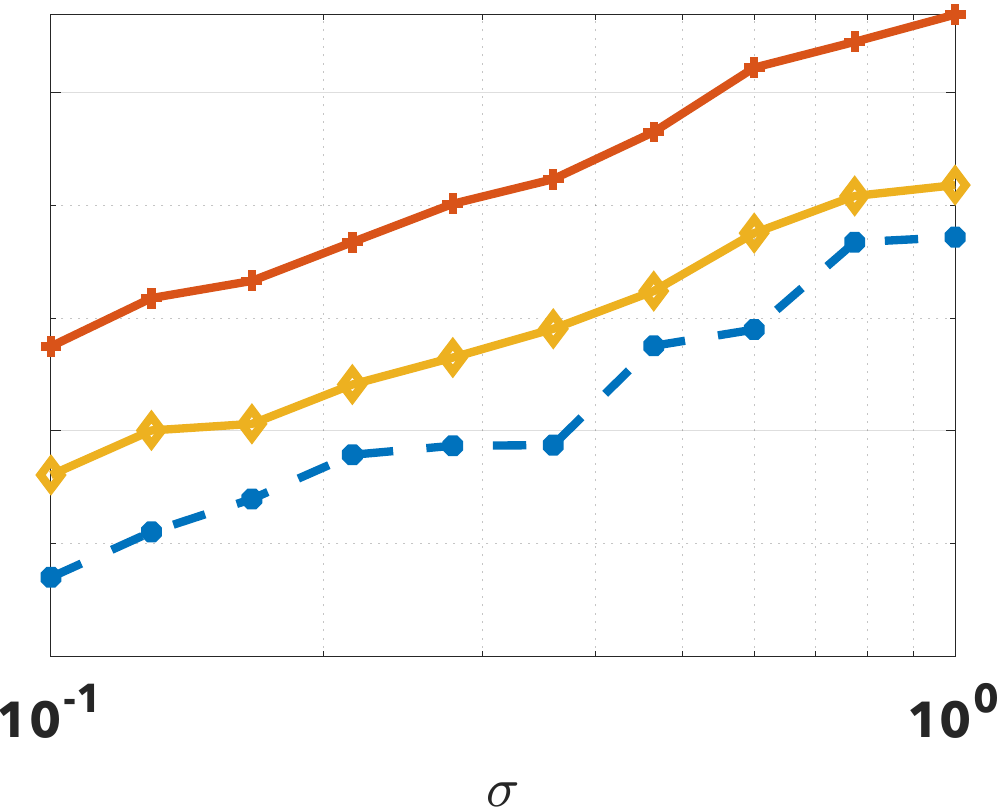}
\includegraphics[height=0.175\textwidth]{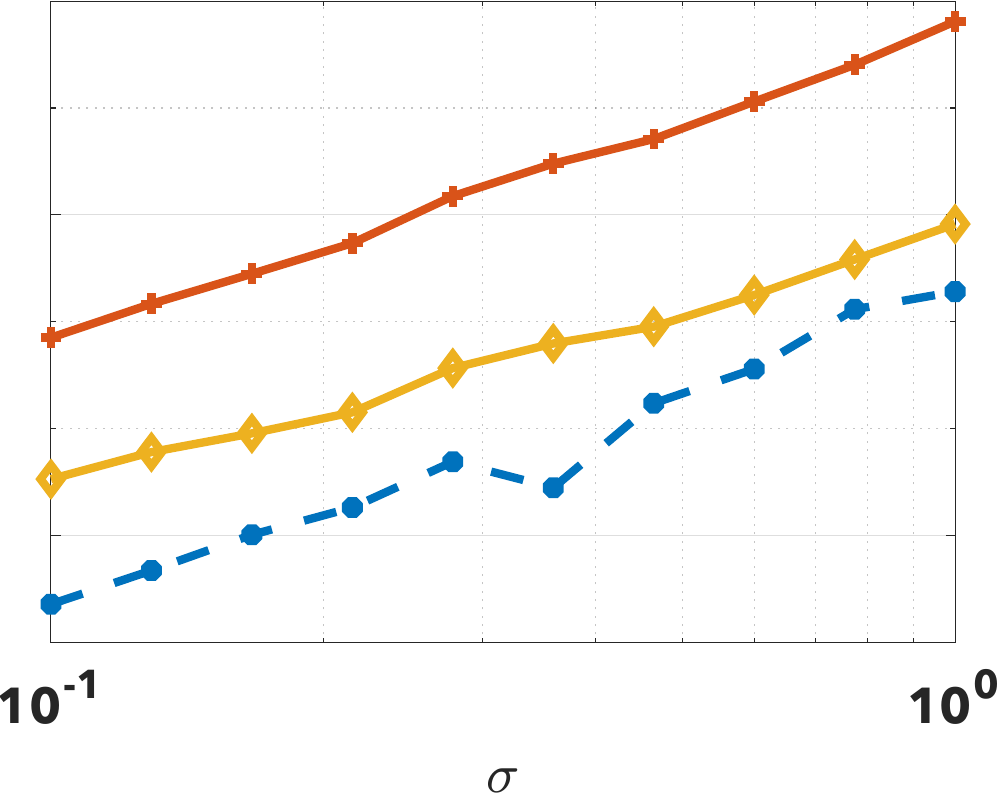}
\includegraphics[height=0.175\textwidth]{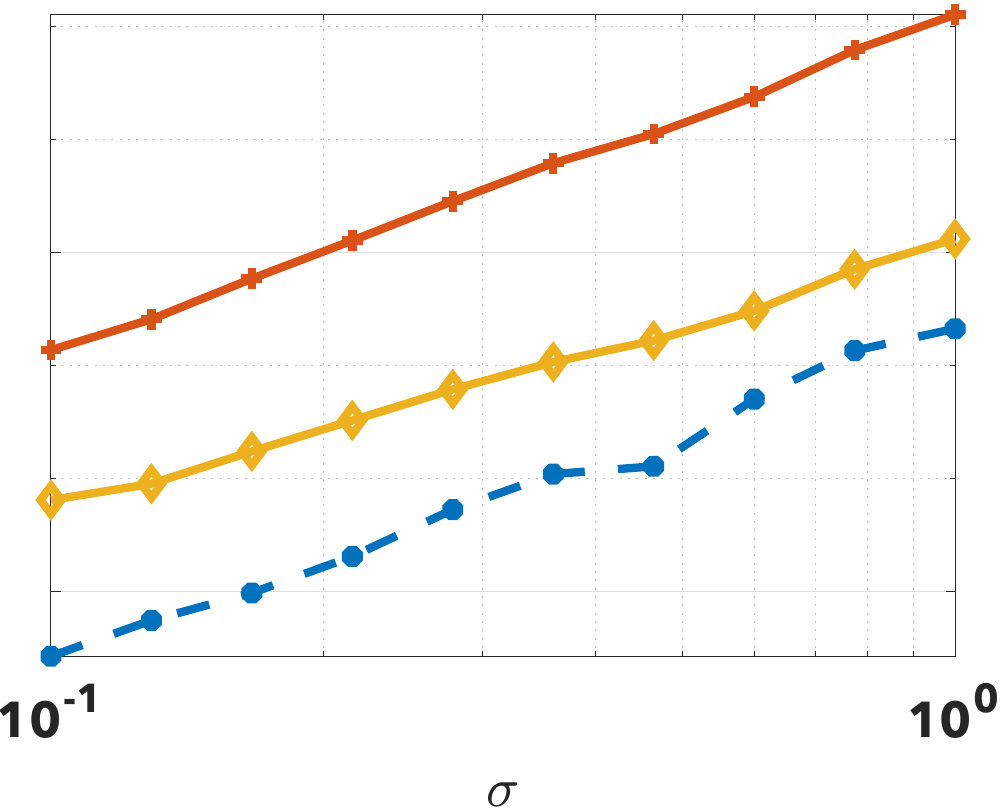}
\includegraphics[height=0.185\textwidth]{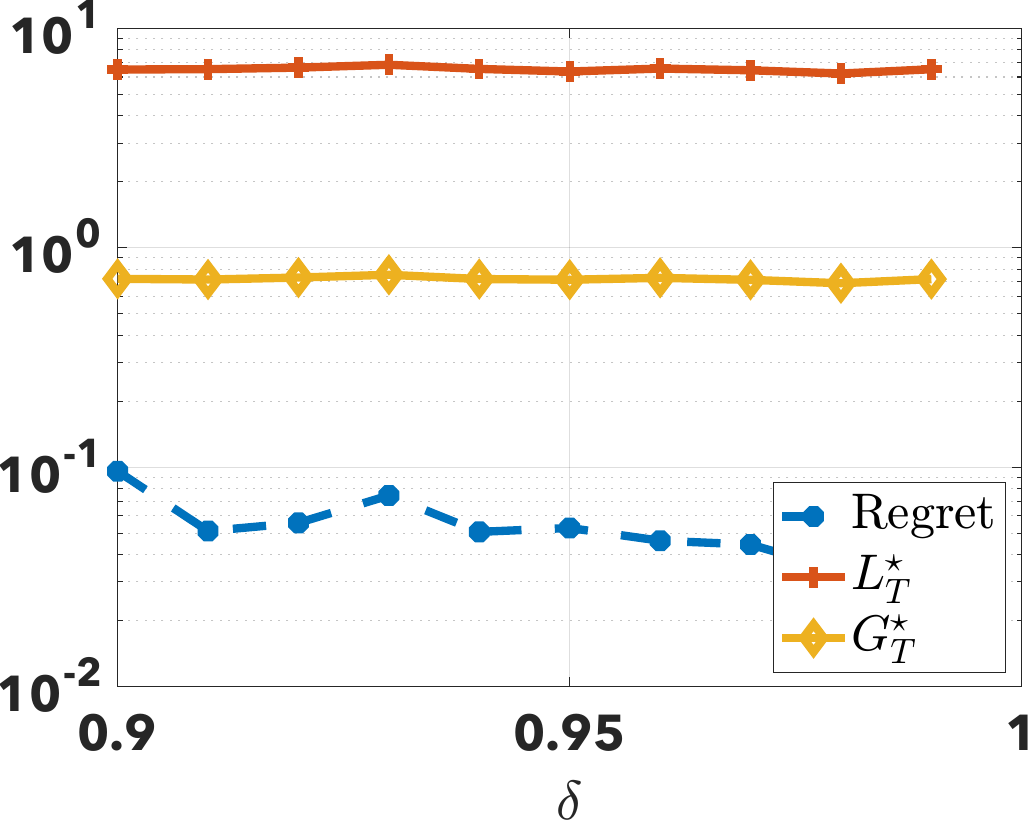}
\includegraphics[height=0.185\textwidth]{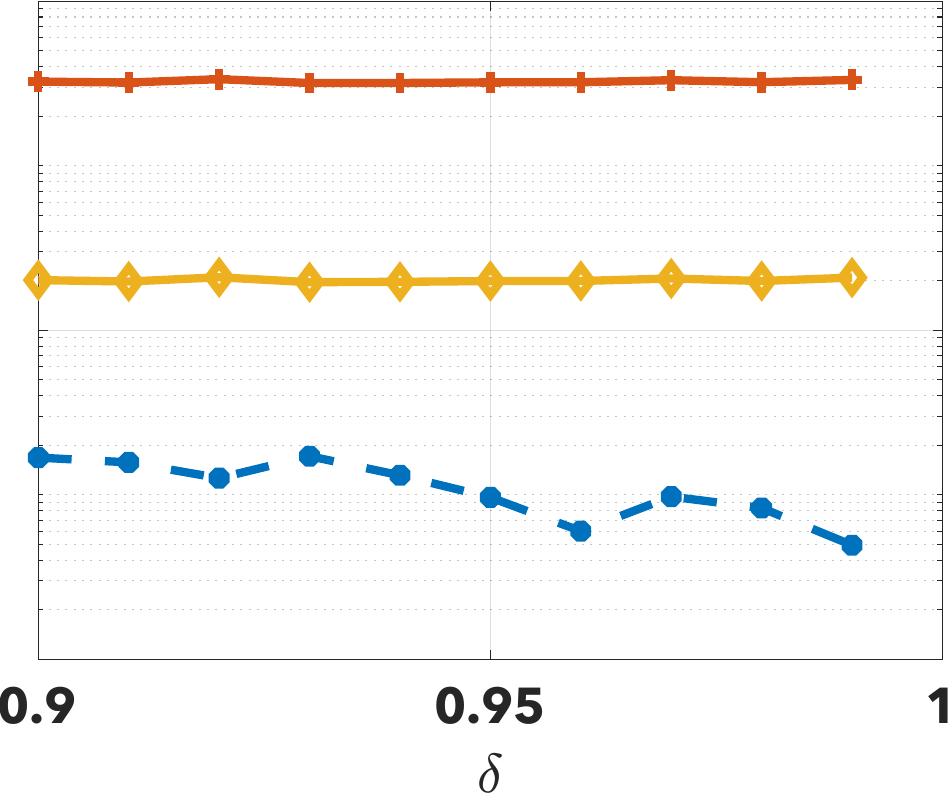}
\includegraphics[height=0.185\textwidth]{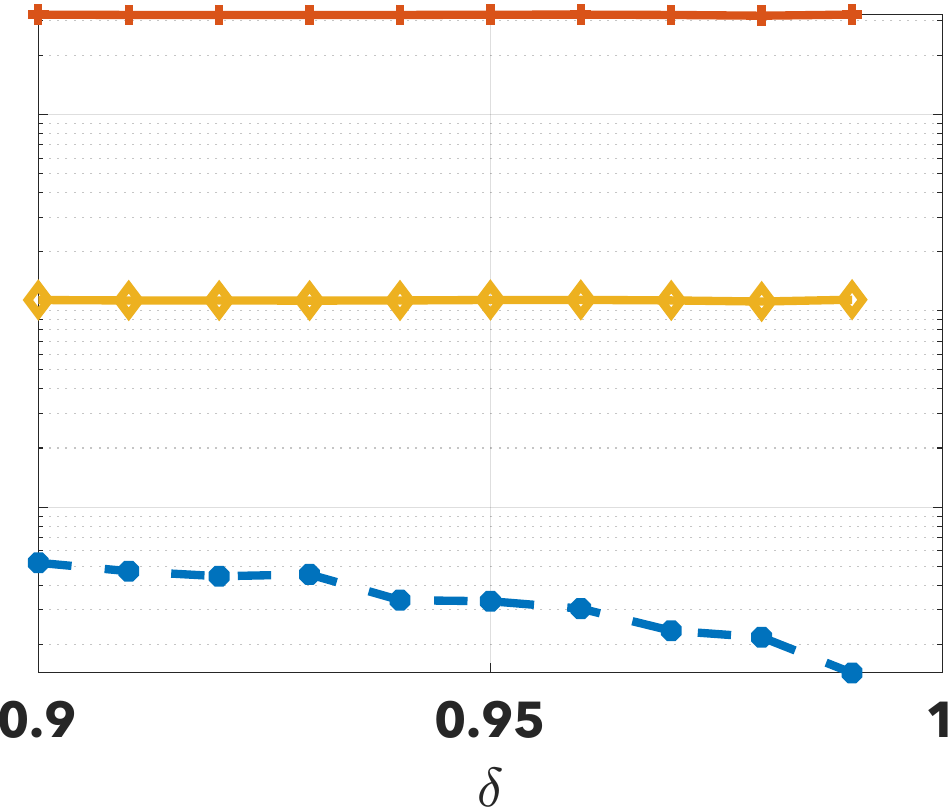}
\includegraphics[height=0.185\textwidth]{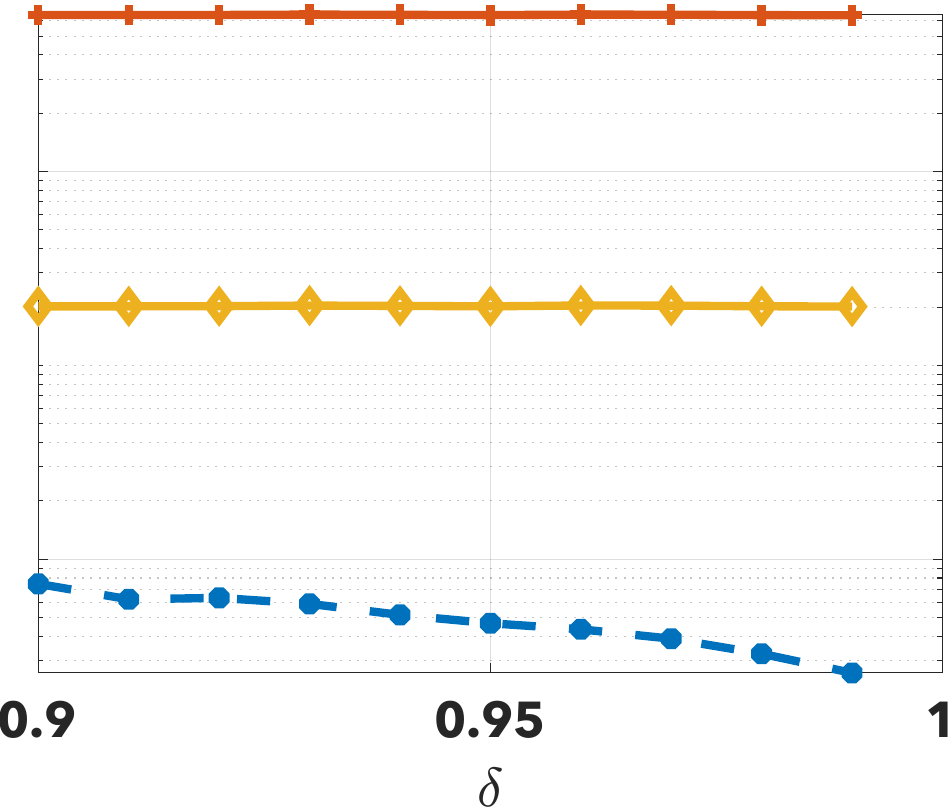}
\caption{Comparison between true regret and the theoretical regret upper bounds given by $L_T^\star$ and $G_T^\star$. First row: experiment on $\ell_p$ regression. From left to right: $T \in \{100, 500, 5000, 10000\}$. \label{fig:lp} Second row: experiment on logistic regression. From left to right: $T \in \{100, 500, 5000, 10000\}$. }
\end{figure}

\Cref{fig:lp} plots 1) the true regret of {\adanorm}; 2) the $G^\star_T$ regret upper bound predicted by \Cref{thm:adanorm-gstar}; 3) the regret upper bound predicted by \Cref{thm:adanorm-gstar} but with $G^\star_T$ replaced by $2L L^\star_T$. The experiment shows that the $G^\star$ regret often yields a substantially tighter characterization of the true algorithm performance, several magnitudes tighter than the $L^\star$-based bound, confirming our theoretical claims.

%% file: sec_conclusion.tex
\section{Conclusion} 
This paper proposes the $G^\star$ regret, a problem-dependent regret that refines the existing $L^\star$ regret. We establish regret upper and lower bounds and extend the $G^\star$ regret to other {\oco} settings. We also showcase the applications of the $G^\star$ regret in stochastic optimization. Experiments validate the efficiency of the $G^\star$ regret.

%% file: sec_app.tex
\section{Proof of results in Section \ref{sec:gstar}}

\subsection{Discussion on Assumption \ref{A1}}

It is known that $L$-smoothness implies \ref{A1} if $\dom \ell_t = \Rbb^n$. Otherwise, \ref{A1} only holds \textit{locally} for points away from the boundary of $\dom \ell_t$ \cite{drori2018properties}.

\begin{prop}[\cite{drori2018properties}, Theorem 3.1] \label{prop:smooth-subset}
Let $\ell: \hat{\Xcal} \rightarrow \Rbb$ be an $L$-smooth and convex function over an open set $\hat{\Xcal} \subsetneq \Rbb^n$. Then for any $x, y \in \hat{\Xcal}$  such that $\|x - y\| \leq \dist(y, \Rbb^n \backslash \hat{\Xcal})$, we have 
\[\ell_t (x) - \ell_t (y) - \langle \nabla \ell_t (y), x - y \rangle \geq \tfrac{1}{2 L} \| \nabla \ell_t (x) - \nabla \ell_t (y) \|^2.\]
\end{prop}

According to \Cref{prop:smooth-subset}, it suffices to take $\Xcal \subsetneq  \dom \ell_t $ such that for all $x, y \in \Xcal$, 
\[\|x - y\| \leq D\leq \min_{z \in \Xcal} ~\dist(z, \Rbb^n \backslash \hat{\Xcal}).\]

\subsection{Algorithms adapting to the $G^\star$ regret}
This section records online algorithms adaptive to the $G^\star$ regret that are used in the main results.

\subsubsection{Online gradient descent (\ogd)}
The algorithm is presented in \Cref{alg:ogd}.

\vspace{-5pt}
\begin{algorithm}[h]
{\textbf{input} initial point $x^1$, learning rate  $\eta > 0$}

\For{$t = 1, 2, \dots$}{
{Play} $x^t$ and receive $\nabla \ell_t (x^t)$

{Update} $x^{t + 1} = \Pi_{\Xcal} [x^t - \eta \nabla \ell_t (x^t)]$\\
}
\caption{Online gradient descent (\ogd) \label{alg:ogd}}
\end{algorithm}
\vspace{-5pt}

\subsubsection{\adanorm} The algorithm is presented in \Cref{alg:adagrad-norm}.
\vspace{-5pt}
\begin{algorithm}[h]
{\textbf{input} initial point $x^1$, learning rate  $\alpha > 0$}

\For{$t = 1, 2, \dots$}{
{Play} $x^t$ and receive $\nabla \ell_t (x^t)$

Update $x^{t + 1} = \Pi_{\Xcal}\Big[x^t - \frac{\alpha}{\sqrt{\sum_{i=1}^t \|\nabla \ell_i (x^i)\|^2}} \nabla \ell_t (x^t)\Big]$ and skip if $\nabla \ell_t (x^t) = 0$\\
}
\caption{\adanorm \label{alg:adagrad-norm}}
\end{algorithm}
\vspace{-5pt}
\subsubsection{\adaftrl} We refer the interested readers to \cite{orabona2018scale} for more details of {\adaftrl}. Define $h (y) = (r (x) +\mathbb{I}
\{ x \in \mathcal{X} \})^{\ast}$, $V_h (x, y) \assign h (x) - h (y) -
\langle \nabla h (y), x - y \rangle$ and $L_t \assign \sum_{j = 1}^t \nabla
\ell_j (x^j)$.
Then recursively define
\[ \textstyle \Delta_t \assign \sum_{j = 1}^t \Delta_{j - 1} V_h ( -
   \tfrac{L_j}{\Delta_{j - 1}}, - \tfrac{L_{j - 1}}{\Delta_{i - 1}} ) \]
and let $\Delta_{j - 1} V_h ( - \tfrac{L_j}{\Delta_{j - 1}}, -
\tfrac{L_{j - 1}}{\Delta_{j - 1}} ) = \lim_{a \rightarrow 0^+}a V_h
( - \tfrac{L_j}{a}, - \tfrac{L_{j - 1}}{a} )$ if $\Delta_{j - 1} =
0$. Finally, define the regularizer 
\[R_t(x) \assign \Delta_t \cdot r(x).\]
The algorithm is given in \Cref{alg:adaftrl}.
\vspace{-5pt}
\begin{algorithm}[h]
{\textbf{input} initial point $x^0$, $\lambda$-strongly convex regularizer $r$}

\For{$t = 1, 2, \dots$}{
Play $x^t = \argmin_{x \in
\mathcal{X}}  \{  \langle L_{t - 1}, x \rangle + R_t
(x) \}$ and receive  $\nabla \ell_t (x^t)$\\
}
\caption{\adaftrl \label{alg:adaftrl}}
\end{algorithm}
\vspace{-5pt}

\subsubsection{Smooth and strongly convex online gradient descent (\texttt{S2OGD})}
The retuned version of \texttt{S2OGD} \cite{wang2020adapting} used in
\Cref{thm:strongly-convex-gstar} is presented in \Cref{alg:s2ogd}.
\vspace{-5pt}
\begin{algorithm}[h]
{\textbf{input} initial point $x^1 \in \Xcal$, smoothness constant $L > 0$,
strong-convexity constant $\mu \in (0, L]$, and gradient-norm bound $M > 0$}

Initialize $S_0 = 0$ and $\Delta = 8 L M^2 / \mu$\\

\For{$t = 1, 2, \dots$}{
Play $x^t$ and receive $g_t = \nabla \ell_t (x^t)$\\

Set $S_t = S_{t - 1} + \| g_t \|^2$ and
$\eta_t = \frac{2 M^2 / \mu}{\Delta + S_t}$\\

Update $x^{t + 1} = \Pi_{\Xcal} [x^t - \eta_t g_t]$\\
}
\caption{Retuned smooth and strongly convex online gradient descent
(\texttt{S2OGD}) \label{alg:s2ogd}}
\end{algorithm}
\vspace{-5pt}

\subsection{Auxiliary results}

\Cref{lem:adanorm} is a well-known result used in the proof of adaptive online algorithms.
\begin{lem}[\cite{orabona2019modern}, Lemma 4.13] \label{lem:adanorm}
  Let $\{ a_t \}_{t = 0}^T$ be a sequence of non-negative numbers and $g$ be
  a non-increasing continuous function, then $\sum_{t = 1}^T a_t g (a_0 + 
  \sum_{j = 1}^t a_j ) \leq \int_{a_0}^{\sum_{t = 1}^T a_t} g (x) ~\mathd x$.
\end{lem}

The following lemmas present the regret guarantees for {\adanorm} and {\adaftrl} applied to linear losses.

\begin{thm}[\cite{orabona2019modern}, Theorem 4.13] \label{thm:auxi-adanorm}
  Suppose \ref{A2} holds. Then the regret of {\adanorm} (\Cref{alg:adagrad-norm}) with $\alpha > 0$ applied to linear losses $\{ g_t \}_{t = 1}^T$
  satisfies $\sum_{t = 1}^T \langle g_t, x^t - x \rangle \leq (\alpha  + \frac{D^2}{2\alpha})
  \sqrt{\sum_{t = 1}^T \| g_t \|^2}$ for any $x \in \mathcal{X}$.
\end{thm}

\begin{thm}[\cite{orabona2018scale}, Theorem 1] \label{thm:auxi-adaftrl}
  Suppose \ref{A2} holds and that $r : \mathcal{X} \rightarrow
  \mathbb{R}_+$ is a non-negative lower semicontinuous function that is
  $\lambda$-strongly convex with respect to Euclidean norm $\| \cdot \|$ and
  is bounded from above. Then the regret of {\adaftrl} (\Cref{alg:adaftrl}) applied to linear losses
  $\{ g_t \}_{t = 1}^T$ satisfies
  \[
    \textstyle \sum_{t = 1}^T \langle g_t, x^t - x \rangle \leq \sqrt{3} \max \{ D, \tfrac{1}{\sqrt{2 \lambda}} \} \sqrt{\sum_{t = 1}^T \| g_t \|^2} (1 + r (x))
  \]
  for any $x \in \mathcal{X}$.
\end{thm}

\subsection{Proof of Proposition \ref{prop:connection}}

We prove different cases separately.

\paragraph{Case 1.} $G_T^{\star} \leq 2 L L^{\star}_T$. To show this
inequality, let $x^{\star} \in \argmin_{x \in \mathcal{X}}  \sum_{t =
1}^T \ell_t (x) $. Using \eqref{eqn:self-bound}, 
\[ \textstyle G^{\star}_T = \sum_{t = 1}^T \| \nabla \ell_t (x^{\star}) \|^2 \leq 2 L
   \sum_{t = 1}^T  [\ell_t (x^{\star}) - \inf \ell_t ]
   = 2 L L^{\star}_T, \]
and this completes the proof.

\paragraph{Case 2.} $G_T^{\star} \ll L^{\star}_T$. To show this inequality, we
adopt \Cref{exp:loss} with $p = \tfrac{1}{6}$ and
\[ G_T^{\star} = \Theta (T^{1 / 3}) \ll \Theta (T^{2 / 3}) = L_T^{\star} . \]
\paragraph{Case 3.} $V_T \ll L^{\star}_T$ and $V_T \ll G^{\star}_T$. To show
this inequality, we adopt the example from Section 6.3 of
{\cite{zhao2024adaptivity}} and consider the following 1D {\oco}
problem with losses $\ell^B_t (x) = \tfrac{1}{2} (a_t x - 1)^2$ and
$\mathcal{X}= [- 1, 1]$, where $a_t = \tfrac{1}{2} - \tfrac{t - 1}{T}$ and $T
= 2 K + 1$ for some integer $K \geq 3$. According to Instance 1 of
{\cite{zhao2024adaptivity}},we have $V_T \leq \sum_{t = 1}^T \tfrac{4}{T^2}
\leq \tfrac{4}{T} = \Ocal(1)$. Next, we consider $G_T^{\star}$ and $L_T^{\star}$. By
definition, we have $\inf \ell^B_t = 0$ and that
\[ \textstyle \sum_{t = 1}^T \ell^B_t (x) = \frac{1}{2} \sum_{t = 1}^T (a_t^2 x^2 - 2 a_t x
   + 1) = \tfrac{1}{2} ( \sum_{t = 1}^T a_t^2 ) x^2 - (
   \sum_{t = 1}^T a_t ) x + \tfrac{T}{2} \]
and the minimizer $x^{\star} = \tfrac{\sum_{t = 1}^T a_t}{\sum_{t = 1}^T
a_t^2} = \tfrac{\sum_{t = 1}^T \frac{1}{2} - \frac{t - 1}{T}}{\sum_{t = 1}^T
( \frac{1}{2} - \frac{t - 1}{T} )^2} = \frac{6 T}{T^2 + 2} \in \Xcal$ gives
\begin{align}
  L_T^{\star} = & \textstyle  \sum_{t = 1}^T \ell^B_t (x^{\star}) = \tfrac{T (T^2 - 1)}{2
  (T^2 + 2)} = \Theta (T) \nonumber\\
  G_T^{\star} = & \textstyle  \sum_{t = 1}^T \| \nabla \ell^B_t (x^{\star}) \|^2 = \tfrac{-
  32 + 60 T^2 - 33 T^4 + 5 T^6}{60 T (2 + T^2)^2} = \Theta (T), \nonumber
\end{align}
where $G_T^{\star} = \Theta (T)$ can be verified by noticing that $\lim_{T
\rightarrow \infty}  \frac{G_T^{\star}}{T} = 12$.

\paragraph{Case 4.} $L_T^{\star} \ll V_T$ and $G_T^{\star} \ll V_T$. To show
this inequality, we again adopt the example from Section 6.3 of
{\cite{zhao2024adaptivity}} and consider the following 1D {\oco}
problem with losses $\ell^C_t (x) = \tfrac{1}{2} (a_t x - b_t)^2$ and
$\mathcal{X}= [- 1, 1]$, where $(a_t, b_t) = (1, 1)$ for even $t$ and $(a_t,
b_t) = ( \tfrac{1}{2}, \tfrac{1}{2} )$ otherwise. According to
Instance 2 of {\cite{zhao2024adaptivity}},we have $V_T = \Theta (T)$. On the
other hand, $x^{\star} = 1$ minimizes each $\ell^C_t$ and $G_T^{\star} =
L_T^{\star} = 0$. This completes the proof.

\subsection{Proof of Lemma \ref{lem:sequence}}
The proof is a special case of \Cref{lem:holder-gstar-conversion} with $\nu = 1$.
%Define $x \assign \sqrt{\sum_{t = 1}^T \| a_t \|^2}$ and $y \assign
%\sqrt{\sum_{t = 1}^T \| b_t \|^2}$. By Cauchy's inequality, we have

%\begin{equation}
% \textstyle \sum_{t = 1}^T \| a_t - b_t \|^2 \geq \sum_{t = 1}^T \| a_t \|^2 + \sum_{t
%  = 1}^T \| b_t \|^2 - 2 \sqrt{\sum_{t = 1}^T \| a_t \|^2} \sqrt{\sum_{t =
%  1}^T \| b_t \|^2} = (x - y)^2 . \nonumber
%\end{equation}
%Then we deduce that
%\begin{align}
% \textstyle \alpha \sqrt{\sum_{t = 1}^T \| a_t \|^2} - \beta \sum_{t = 1}^T \| a_t - b_t
%  \|^2 \leq{} & \alpha x - \beta (x - y)^2 \nonumber\\
%  \leq{} & \max_x ~\{ \alpha x - \beta (x - y)^2 \} \nonumber\\
%  ={} & \alpha y + \tfrac{\alpha^2}{4 \beta} \nonumber\\
%  ={} & \textstyle \alpha \sqrt{\sum_{t = 1}^T \| b_t \|^2} + \tfrac{\alpha^2}{4 \beta},
%  \nonumber
%\end{align}
%and this completes the proof.

\subsection{Proof of Theorem \ref{thm:adanorm-gstar}}
Using \Cref{thm:auxi-adanorm} with $\alpha=\frac{\sqrt{2}D}{2}$and \ref{A2}, we deduce that
\begin{align}
  \rho_T(x) ={} & \textstyle  \sum_{t = 1}^T \ell_t (x^t) - \ell_t (x) \nonumber\\
  \leq{} & \textstyle \sum_{t=1}^T \langle \nabla\ell_t(x^t), x^t - x\rangle -
  \tfrac{1}{2 L} \sum_{t = 1}^T \| \nabla \ell_t (x^t) - \nabla \ell_t (x)
  \|^2 \label{eqn:adanorm-gstar-1} \\
  \leq{} & \textstyle  \sqrt{2} D \sqrt{\sum_{t = 1}^T \| \nabla \ell_t (x^t) \|^2} -
  \tfrac{1}{2 L} \sum_{t = 1}^T \| \nabla \ell_t (x^t) - \nabla \ell_t (x)
  \|^2 \label{eqn:adanorm-gstar-2} \\
  \leq{} & \textstyle \sqrt{2}D \sqrt{\sum_{t = 1}^T \| \nabla \ell_t (x) \|^2} + L D^2, \label{eqn:adanorm-gstar-3}
\end{align}
where \eqref{eqn:adanorm-gstar-1} applies \ref{A1}, \eqref{eqn:adanorm-gstar-2} uses \Cref{thm:auxi-adanorm}, and 
\eqref{eqn:adanorm-gstar-3} uses \Cref{lem:sequence} with $a_t = \nabla \ell_t(x^t),b_t = \nabla \ell_t(x), \alpha = \sqrt{2}D$, and $\beta = \frac{1}{2L}$. This completes the proof.

\subsection{Proof of Theorem \ref{thm:adaftrl-gstar}}

Recall that $\hat{\ell}_t (x) \assign \langle \nabla \ell_t (x^t), x \rangle$.
Consider the online linear optimization problem minimizing $\sum_{t = 1}^T \hat{\ell}_t (x^t) - \hat{\ell}_t (x)$ for any fixed decision $x \in \Xcal$ in hindsight.
The choice $\lambda \geq \frac{1}{2D^2}$ ensures that $\max \{
     D, \tfrac{1}{\sqrt{2 \lambda}} \} = D$, and with \Cref{thm:auxi-adaftrl}, {\adaftrl} satisfies
\begin{equation} \label{eqn:ada-ftrl-0}
	\textstyle  \sum_{t = 1}^T \hat{\ell}_t (x^t) - \hat{\ell}_t (x) \leq \sqrt{3} D
   \sqrt{\sum_{t = 1}^T \| \nabla \hat{\ell}_t (x^t) \|^2} (1 + r (x)) \leq
   \sqrt{3} R D \sqrt{\sum_{t = 1}^T \| \nabla \hat{\ell}_t (x^t) \|^2},
\end{equation}
   where the last inequality applies the definition $R\assign \max_{x\in\Xcal} r(x)+1$. Using \ref{A1} and \ref{A2}, we deduce that
\begin{align}
  \rho_T(x) ={} & \textstyle \sum_{t = 1}^T \ell_t (x^t) - \ell_t (x) \nonumber\\
  \leq{} & \textstyle \sum_{t = 1}^T [ \langle \nabla \ell_t (x^t), x^t - x \rangle -
  \tfrac{1}{2 L} \| \nabla \ell_t (x^t) - \nabla \ell_t (x) \|^2 ]
  \nonumber\\
  ={} & \textstyle \sum_{t = 1}^T [ \hat{\ell}_t (x^t) - \hat{\ell}_t (x) -
  \tfrac{1}{2 L} \| \nabla \ell_t (x^t) - \nabla \ell_t (x) \|^2 ]
  \nonumber\\
  \leq{} & \textstyle \sqrt{3} R D \sqrt{\sum_{t = 1}^T \| \nabla \hat{\ell}_t (x^t) \|^2}
  - \tfrac{1}{2 L} \sum_{t = 1}^T \| \nabla \ell_t (x^t) - \nabla \ell_t (x)
  \|^2 \label{eqn:ada-ftrl-0-1}\\
  ={} & \textstyle \sqrt{3} R D \sqrt{\sum_{t = 1}^T \| \nabla \ell_t (x^t) \|^2} -
  \tfrac{1}{2 L} \sum_{t = 1}^T \| \nabla \ell_t (x^t) - \nabla \ell_t (x)
  \|^2, \label{eqn:adaftrl-1} \\
  \leq{} &\textstyle \sqrt{3} R D \sqrt{\sum_{t = 1}^T \| \nabla \ell_t (x) \|^2} + 2 R^2 L D^2, \label{eqn:adaftrl-2}
\end{align}
where \eqref{eqn:ada-ftrl-0-1} plugs in \eqref{eqn:ada-ftrl-0}; \eqref{eqn:adaftrl-1} uses the fact $\nabla \hat{\ell}_t (x^t) = \nabla
\ell_t (x^t)$ and \eqref{eqn:adaftrl-2} again applies \Cref{lem:sequence} with $a_t = \nabla \ell_t(x^t),b_t = \nabla \ell_t(x), \alpha = \sqrt{3}RD$, and $\beta = \frac{1}{2L}$. This completes the proof.

\subsection{Proof of Theorem \ref{thm:lb}}

We use the standard online linear optimization lower bound construction with Rademacher random variables; see, for example, Theorem 5.1 of \cite{orabona2019modern}. Let $u, v \in \Xcal$ satisfy $\| u - v \| = D$, and define $q \assign (u - v) / D$. Then $\| q \| = 1$ and
\[
  \textstyle \max_{x \in \Xcal} \langle q, x \rangle - \min_{x \in \Xcal} \langle q, x \rangle = D.
\]
Write $a \assign \min_{x \in \Xcal} \langle q, x \rangle$, $b \assign \max_{x \in \Xcal} \langle q, x \rangle$, and $c \assign (a + b) / 2$. Let $\{ \sigma_t \}_{t = 1}^T$ be independent Rademacher random variables and define
\[
  \textstyle g_t \assign M \sigma_t q \quad \text{and} \quad \ell_t (x) \assign \langle g_t, x \rangle.
\]
Let $S_T \assign \sum_{t = 1}^T \sigma_t$. Since $x^t$ depends only on the previous losses and the internal randomization of the algorithm, $\sigma_t$ is independent of $x^t$. Moreover,
\[
  \textstyle \min_{x \in \Xcal} \sum_{t = 1}^T \ell_t (x) = M S_T c - \tfrac{M D}{2} |S_T|.
\]
Consequently,
\[
  \textstyle \rho_T (x^\star) = M \sum_{t = 1}^T \sigma_t [\langle q, x^t \rangle - c] + \tfrac{M D}{2} |S_T|.
\]
Taking expectation over the Rademacher variables and the internal randomization of the algorithm, and applying Khintchine's inequality, gives
\[
  \textstyle \mathbb{E} [\rho_T (x^\star)] = \tfrac{M D}{2} \mathbb{E} [|S_T|] \geq \tfrac{M D}{2} \sqrt{\tfrac{T}{2}} \geq \tfrac{M D \sqrt{T}}{4}.
\]
For every realization, $\nabla \ell_t (x) = g_t$ for all $x$, and hence
\[
  \textstyle G_T^\star = \sum_{t = 1}^T \| g_t \|^2 = M^2 T.
\]
Averaging over the Rademacher variables therefore shows that there exists a deterministic realization of the loss sequence satisfying
\[
  \textstyle \mathbb{E} [\rho_T (x^\star)] \geq \tfrac{D}{4} \sqrt{G_T^\star},
\]
where the remaining expectation is only over the internal randomization of the algorithm. For a deterministic algorithm, this expectation can be omitted. Finally, linear losses satisfy \ref{A1} for every $L > 0$, which completes the proof.

\section{Proof of results in Section \ref{sec:extension}}

%\subsection{Proof of Theorem \ref{thm:dynamic-gstar}}

\subsection{Dynamic regret and proof of results in Section \ref{sec:dynamic}}

This section establishes an $\mathcal{O} ( \sqrt{(1 +2\mathcal{P}_T +
\hat{G}_T) (1 +2\mathcal{P}_T)} \log \log T)$ regret upper bound using the meta-expert algorithm framework developed in
{\cite{zhao2024adaptivity,zhao2020dynamic}}. For brevity, this paper focuses on $\mathrm{Sword}_{\mathrm{small}}$, a simpler version of the framework developed in {\cite{zhao2020dynamic}}. Our analyses are compatible with the more
sophisticated version in {\cite{zhao2024adaptivity}}. Following the setup in
{\cite{zhao2020dynamic}}, our analysis resorts to an extra constant
\[ M \assign \max_{t \in [T]} ~\max_{x \in \mathcal{X}}  \| \nabla \ell_t (x)
   \|, \]
which can be further bounded by taking any $\bar{x} \in \mathcal{X}$ and
$\max_{t \in [T]} \max_{x \in \mathcal{X}}  \| \nabla \ell_t (x) \| \leq
\max_{t \in [T]}  \| \nabla \ell_t (\bar{x}) \| + L D$. This constant $M$ will appear in a $\log \log$ term in the final result.

\subsubsection{Algorithm design}

$\mathrm{Sword}_{\mathrm{small}}$
adopts \texttt{VanillaHedge} (exponentiated gradient) of a meta algorithm to choose the learning rate for the lower-level expert
algorithm (\ogd, \Cref{alg:ogd}).
 The meta algorithm operates on a grid of learning rates defined as follows
\[ \mathcal{N}= \{ \eta : \eta = 2^k \cdot \eta_{\min}, k \in [N] \} \]
where 
\begin{equation} \label{eqn:grid-param}
	\eta_{\min} \assign \tfrac{D}{M} \sqrt{\tfrac{1}{2 T}} \qquad\text{and}\qquad N \assign
\Big\lceil \tfrac{1}{\log 2} \log ( \tfrac{M \sqrt{2 T}}{L D})
\Big\rceil.
\end{equation}
Then, the exponentiated gradient combines the iterates from $N$
copies of {\ogd} with different learning rates. The whole algorithm is presented in \Cref{alg:sword}.
\begin{algorithm}[h]
{\textbf{input} initial vector $p^1 = \tfrac{1}{N} \textbf{1}_N$ and $\eta_i = \eta_{\min} 2^{i - 1},
  i \in [N]$, expotentiated gradient learning rate $\delta > 0$, initial point $x^1_i = x^1, i\in[N]$.}
  
\For{$t = 1, 2, \dots$}{
Play $x^t = \sum_{i=1}^N p_i^t x_i^t $ and receive $\nabla \ell_t (x^t)$.

Update $x^{t + 1}_i = \Pi_\Xcal [x^t_i - \eta_i \nabla \ell_t (x^t)], i\in[N]$\\
Update $p^{t + 1}_i \propto p^t_i \exp (- \delta \langle \nabla \ell_t (x^t),
  x^t_i \rangle), i \in [N]$\\
}
\caption{$\mathrm{Sword}_{\mathrm{small}}$ \cite{zhao2020dynamic} \label{alg:sword}}
\end{algorithm}
\vspace{-5pt}

\subsubsection{Auxiliary results}

Before we proceed, we present several auxiliary results that tighten the existing analyses through \ref{A1}.
\begin{lem}[Meta-regret]
  \label{lem:meta}Under \ref{A1} and \ref{A2}, \Cref{alg:sword} satisfies
  \[ \textstyle \sum_{t = 1}^T \ell_t (x^t) - \sum_{t = 1}^T \ell_t (x^t_i) \leq \textstyle \tfrac{2 + \log N}{\delta} + \delta D^2 \sum_{t = 1}^T \| \nabla
  \ell_t (x^t) \|^2 -
     \tfrac{1}{2 L} \sum_{t = 1}^T \| \nabla \ell_t (x^t) - \nabla \ell_t
     (x^t_i) \|^2 \]
  for any $i \in [N]$, where $N$ is defined in \eqref{eqn:grid-param}.
\end{lem}

\begin{proof}
By \ref{A1}, we have, for any $i \in [N]$, that
\begin{align}
 \textstyle \sum_{t = 1}^T \ell_t (x^t) - \sum_{t = 1}^T \ell_t (x^t_i) \leq  \sum_{t =
  1}^T \langle \nabla \ell_t (x^t), x^t - x^t_i \rangle - \tfrac{1}{2 L} \|
  \nabla \ell_t (x^t) - \nabla \ell_t (x^t_i) \|^2 . \label{eqn:meta-1}
\end{align}
The following proof is adapted from the proof of Theorem 11 of
{\cite{zhao2020dynamic}}. Using the regret guarantees of exponentiated
gradient applied to linearized loss $\langle \nabla \ell_t (x^t), x \rangle$, we have, by Theorem 19 of \cite{syrgkanis2015fast}, that,
\begin{align}
 \textstyle \sum_{t = 1}^T \langle \nabla \ell_t (x^t), x^t - x^t_i \rangle \leq{} & \textstyle
  \tfrac{2 + \log N}{\delta} + \delta \sum_{t = 1}^T \max_{i \in [N]} \langle
  \nabla \ell_t (x^t), x_i^t \rangle^2 \nonumber\\
  \leq{} & \textstyle \tfrac{2 + \log N}{\delta} + \delta D^2 \sum_{t = 1}^T \| \nabla
  \ell_t (x^t) \|^2 \label{eqn:meta-2}
\end{align}

combining \eqref{eqn:meta-1} with \eqref{eqn:meta-2} completes the proof.
\end{proof}

The next result bounds the expert regret.

\begin{lem}[Expert regret]
  \label{lem:expert}Under \ref{A1} and \ref{A2} and suppose $\eta \in ( 0, \tfrac{1}{4 L}
  ]$. Then
  \[\textstyle\hat{\rho} (\{ \hat{x}_t \}_{t = 1}^T) \leq \tfrac{D (D
     +2\mathcal{P}_T)}{2 \eta} + \eta \hat{G}_T - \tfrac{1}{4 L} \sum_{t = 1}^T
     \| \nabla \ell_t (x^t_i) - \nabla \ell_t (\hat{x}_t) \|^2 . \]
for all $i \in [N]$.
\end{lem}

\begin{proof}
First, we deduce that
\begin{align}
  \| x^{t + 1} - \hat{x}_t \|^2 \leq{} & \| x^t - \eta \nabla \ell_t (x^t) -
  \hat{x}_t \|^2 \nonumber\\
  ={} & \| x^t - \hat{x}_t \|^2 - 2 \eta \langle \nabla \ell_t (x^t), x^t -
  \hat{x}_t \rangle + \eta^2 \| \nabla \ell_t (x^t) \|^2 \nonumber\\
  \leq{} & \| x^t - \hat{x}_t \|^2 - 2 \eta [\ell_t (x^t) - \ell_t (\hat{x}_t)]
  \underbrace{- \tfrac{\eta}{L} \| \nabla \ell_t (x^t) - \nabla \ell_t
  (\hat{x}_t) \|^2 + \eta^2 \| \nabla \ell_t (x^t) \|^2}_{{\heartsuit}}.
  \nonumber
\end{align}
Next, we upper bound $\heartsuit$ for $\eta \in ( 0, \tfrac{1}{L} )$ with Young's inequality $\|a + b\|^2 \leq (1 + \gamma)\|a\|^2 +(1 + \gamma^{-1})\|b\|^2$:
\begin{align}
  & - \tfrac{\eta}{L} \| \nabla \ell_t (x^t) - \nabla \ell_t (\hat{x}_t) \|^2
  + \eta^2 \| \nabla \ell_t (x^t) \|^2 \nonumber\\
  ={} & - \tfrac{\eta}{L} \| \nabla \ell_t (x^t) - \nabla \ell_t (\hat{x}_t) \|^2
  + \eta^2 \| \nabla \ell_t (x^t) - \nabla \ell_t (\hat{x}_t)+\nabla \ell_t (\hat{x}_t) \|^2 \nonumber\\
  \leq{} & \eta [(1+\gamma)\eta - \tfrac{1}{L}]  \| \nabla \ell_t (x^t) - \nabla \ell_t (\hat{x}_t) \|^2+ (1 + \gamma^{-1})\eta^2 \|\nabla \ell_t (\hat{x}_t) \|^2. \label{eqn:dynamic-regret-1}
\end{align}

For each $i \in [N]$, by \eqref{eqn:dynamic-regret-1}, we have $ \eta (1 + \gamma) - \tfrac{1}{L} = 2\eta -
\tfrac{1}{L} \leq - \tfrac{1}{2 L}$ when $\eta \leq \tfrac{1}{4 L}$ and $\gamma = 1$. Hence
\[\| x^{t + 1} - \hat{x}_t \|^2 \leq \| x^t - \hat{x}_t \|^2 - 2 \eta [\ell_t (x^t) - \ell_t (\hat{x}_t)] + 2 \eta^2\| \nabla
\ell_t (\hat{x}_t) \|^2 - \tfrac{\eta}{2L}\| \nabla \ell_t (x^t_i) - \nabla \ell_t (\hat{x}_t) \|^2.\] 
Instead of dropping $\| \nabla \ell_t
(x^t_i) - \nabla \ell_t (\hat{x}_t) \|^2$, keep it throughout the standard dynamical regret analysis \cite{zhang2019adaptive} completes the proof.
\end{proof}

\subsubsection{Algorithm analysis}

\begin{thm}[Formal version of \Cref{thm:dynamic-gstar-adaptive}] \label{thm:dynamic-gstar-formal}
  Under \ref{A1}, \ref{A2} and letting $\delta =
  \sqrt{\tfrac{2 + \log N}{D^2 \sum_{t = 1}^T \| \nabla \ell_t (x^t) \|^2}}$, \Cref{alg:sword} satisfies
  \[ \textstyle \hat{\rho} (\{ \hat{x}_t \}_{t = 1}^T) \leq 8 \sqrt{(3 + \log N) {D^2}  +2\mathcal{P}_T D} + 8 (3 + \log N) L D^2 + 8LD\Pcal_T  = \mathcal{O} ( \sqrt{(1 +\mathcal{P}_T + \hat{G}_T) (1 +\mathcal{P}_T)} \log \log T)\]
  for any comparator sequence $\{ \hat{x}_t \}_{t = 1}^T, \hat{x}_t \in
  \mathcal{X}$, where $N = \big\lceil \tfrac{1}{\log 2} \log ( \tfrac{M \sqrt{2 T}}{L D}) \big\rceil =\Ocal(\log T)$ is given by \eqref{eqn:grid-param}.
\end{thm}

\begin{proof}
Define $\eta^{\star} \assign \min \{ \sqrt{\tfrac{D (D +2\mathcal{P}_T)}{2
\hat{G}_T}}, \tfrac{1}{4L} \}$. We start by the standard  case where $\eta^\star = \sqrt{\tfrac{D (D +2\mathcal{P}_T)}{2
\hat{G}_T}}$.  Since $\hat{G}_T = \sum_{t = 1}^T \|
\nabla \ell_t (\hat{x}^t) \|^2 \leq M^2 T$, we have
\[ \eta_{\min} = \tfrac{D}{M} \sqrt{\tfrac{1}{2 T}} \leq \eta^{\star} \leq
   \tfrac{1}{4 L} = \eta_{\max} \]
By construction, there must exist some $k \in \mathcal{N}$ such that $\eta_k =
2^k \cdot \eta_{\min}$, $\eta_k \leq \eta^{\star} \leq 2 \eta_k$, and by \Cref{lem:expert},
\begin{equation} \label{eqn:dynamic-gstar-1}
\textstyle \sum_{t = 1}^T \ell_t (x^t_k) - \sum_{t = 1}^T \ell_t (\hat{x}_t) \leq 4
   \sqrt{D (D +2\mathcal{P}_T) \hat{G}_T} - \tfrac{1}{4 L} \sum_{t = 1}^T \|
   \nabla \ell_t (x^t_k) - \nabla \ell_t (\hat{x}_t) \|^2 .
\end{equation}
Combining \Cref{lem:meta} and \eqref{eqn:dynamic-gstar-1}, we deduce that
\begin{align}
  \hat{\rho} (\{ \hat{x}_t \}_{t = 1}^T) ={} & \textstyle \sum_{t = 1}^T \ell_t (x^t) -
  \sum_{t = 1}^T \ell_t (\hat{x}_t) \label{eqn:dynamic-gstar-start}\\
  ={} & \textstyle \sum_{t = 1}^T \ell_t (x^t) - \sum_{t = 1}^T \ell_t (x^t_k) + \sum_{t =
  1}^T \ell_t (x^t_k) - \sum_{t = 1}^T \ell_t (\hat{x}_t) \nonumber\\
  \leq{} & \textstyle \tfrac{2 + \log N}{\delta} + \delta D^2 \sum_{t = 1}^T \| \nabla
  \ell_t (x^t) \|^2 + 4 \sqrt{D (D +2\mathcal{P}_T) \hat{G}_T} \nonumber\\
  & \textstyle- \tfrac{1}{2 L} \sum_{t = 1}^T \| \nabla \ell_t (x^t_k) - \nabla \ell_t
  (\hat{x}_t) \|^2 - \tfrac{1}{4 L} \sum_{t = 1}^T \| \nabla \ell_t (x^t) -
  \nabla \ell_t (x^t_k) \|^2   \label{eqn:dynamic-gstar-2}\\
  \leq{} & \textstyle \tfrac{2 + \log N}{\delta} + \delta D^2 \sum_{t = 1}^T \| \nabla
  \ell_t (x^t) \|^2 + \tfrac{D (D +2\mathcal{P}_T)}{2 \eta_k} + \eta_k
  \hat{G}_T \nonumber\\
  & \textstyle - \tfrac{1}{4 L} \sum_{t = 1}^T [\| \nabla \ell_t (x^t_k) - \nabla \ell_t
  (\hat{x}_t) \|^2 + \| \nabla \ell_t (x^t) - \nabla \ell_t (x^t_k) \|^2]
\nonumber \\
  \leq{} & \textstyle \tfrac{2 + \log N}{\delta} + \delta D^2 \sum_{t = 1}^T \| \nabla
  \ell_t (x^t) \|^2 + 4 \sqrt{D (D +2\mathcal{P}_T) \hat{G}_T} - \tfrac{1}{8 L}
  \sum_{t = 1}^T \| \nabla \ell_t (x^t) - \nabla \ell_t (\hat{x}_t) \|^2,
  \label{eqn:dynamic-gstar-3}
\end{align}
where  \eqref{eqn:dynamic-gstar-2} plugs in the meta regret \Cref{lem:meta} and expert regret \Cref{lem:expert}; \eqref{eqn:dynamic-gstar-3} uses $\|a\|^2 + \|b\|^2 \geq \frac{1}{2}\|a + b\|^2$. Finally, taking $\delta = \sqrt{\tfrac{2 + \log N}{D^2 \sum_{t = 1}^T \| \nabla \ell_t
(x^t) \|^2}}$, gives
\[ \textstyle \hat{\rho} (\{ \hat{x}_t \}_{t = 1}^T) \leq 2 D \sqrt{(2 + \log N) \sum_{t
   = 1}^T \| \nabla \ell_t (x^t) \|^2} - \tfrac{1}{8 L} \sum_{t = 1}^T \|
   \nabla \ell_t (x^t) - \nabla \ell_t (\hat{x}_t) \|^2 + 4 \sqrt{D (D
   +2\mathcal{P}_T) \hat{G}_T}. \]
Invoking \Cref{lem:sequence}, we have 
\begin{align}
 \textstyle \hat{\rho}(\{ \hat{x}_t \}_{t = 1}^T)\leq{} & 2 D \textstyle\sqrt{(2 + \log N) \sum_{t = 1}^T \| \nabla \ell_t (x^t) \|^2} -
  \tfrac{1}{8 L} \sum_{t = 1}^T \| \nabla \ell_t (x^t) - \nabla \ell_t
  (\hat{x}_t) \|^2 + 4 \sqrt{D (D +2\mathcal{P}_T) \hat{G}_T} \nonumber\\
  \leq{} & \textstyle ( 2 \sqrt{(2 + \log N) D^2} + 4 \sqrt{D (D +2\mathcal{P}_T)}
  ) \sqrt{\hat{G}_T} + 8 (2 + \log N) L D^2 \nonumber\\
  \leq{} & 8 \sqrt{(3 + \log N) {D^2}  +2\mathcal{P}_T D} {\textstyle\sqrt{\hat{G}_T}} + 8 (2 + \log N) L D^2, \label{eqn:dynamic-gstar-4}
\end{align}
where \eqref{eqn:dynamic-gstar-4} uses $\sqrt{x} + \sqrt{y} \leq 2\sqrt{x + y}$ and this completes the proof when $\eta^\star = \sqrt{\tfrac{D (D +2\mathcal{P}_T)}{2 \hat{G}_T}}$. 

Finally, we consider the case where $\eta^\star = \frac{1}{4L}$, in which case $\hat{G}_T \leq 8 L^2 D(D+2 \Pcal_T)$ and for $\eta_k = \eta_{\max}$, we have

\[\textstyle \sum_{t = 1}^T \ell_t (x^t_k) - \sum_{t = 1}^T \ell_t (\hat{x}_t) \leq 4 L D(D + 2 \Pcal_T) - \tfrac{1}{4 L} \sum_{t = 1}^T \|
   \nabla \ell_t (x^t_k) - \nabla \ell_t (\hat{x}_t) \|^2.\]
   
Repeating the same regret analysis from \eqref{eqn:dynamic-gstar-start} to \eqref{eqn:dynamic-gstar-3} gives 
\begin{align}
 \textstyle \hat{\rho}(\{ \hat{x}_t \}_{t = 1}^T)\leq{} & 2 D \textstyle\sqrt{(2 + \log N) \sum_{t = 1}^T \| \nabla \ell_t (x^t) \|^2} -
  \tfrac{1}{8 L} \sum_{t = 1}^T \| \nabla \ell_t (x^t) - \nabla \ell_t
  (\hat{x}_t) \|^2 + 4LD(D+2\Pcal_T) \nonumber\\
  \leq{} & \textstyle 2 \sqrt{(2 + \log N) D^2}  \sqrt{\hat{G}_T} + 8 (2 + \log N) L D^2 + 4LD(D+2\Pcal_T) \label{eqn:dynamic-gstar-5}
\end{align}
Combining \eqref{eqn:dynamic-gstar-4} and \eqref{eqn:dynamic-gstar-5} completes the proof.
\end{proof}

\begin{rem}
  As mentioned in {\cite{zhao2020dynamic}} the choice of $\delta$ in the meta algorithm still requires knowledge of unknown quantify $\sum_{t = 1}^T \| \nabla \ell_t
  (x^t) \|^2$. This issue can be addressed using the doubling trick or an adaptive
  version of the hedge algorithm discussed in Appendix B of {\cite{zhao2024adaptivity}} or Section 7.6 of \cite{orabona2019modern}.
\end{rem}

\subsection{Bandit feedback and proof of results in Section \ref{sec:bandit}}
This section studies online algorithms adaptive to the $G^\star$ regret under bandit feedback. Specifically, we focus on \bgd~with a constant learning rate and \bgd~with an \adanorm~learning rate. The details of both algorithms are provided below.

\vspace{-5pt}
\begin{algorithm}[h]
{\textbf{input} initial point $x^1$, learning rate  $\eta > 0$}

\For{$t = 1, 2, \dots$}{
{Draw} a random direction $s_t$ uniformly from the unit sphere $\Sbb_n$

{Query} $(y^t_{+}, y^t_{-}) = ( x^t + \mu s_t, x^t - \mu s_t)$ and receive $\ell_t(y^t_{+})$, $\ell_t(y^t_{-})$

{Construct} $g_t = \tfrac{n}{2\mu}\bigl[\ell_t(y^t_{+}) - \ell_t(y^t_{-})\bigr] s_t$

{Update} $x^{t + 1} = \Pi_{\Xcal} [x^t - \eta g_t]$\\
}
\caption{Bandit gradient descent (\bgd) \label{alg:bgd}}
\end{algorithm}
\vspace{-5pt}
\begin{algorithm}[h]
{\textbf{input} initial point $x^1$, learning rate  $\alpha > 0$, smoothing parameter $\mu > 0$}

\For{$t = 1, 2, \dots$}{
{Draw} a random direction $s_t$ uniformly from the unit sphere $\Sbb_n$

{Query} $(y^t_{+}, y^t_{-}) = ( x^t + \mu s_t, x^t - \mu s_t)$ and receive $\ell_t(y^t_{+})$, $\ell_t(y^t_{-})$

{Construct} $g_t = \tfrac{n}{2\mu}\bigl[\ell_t(y^t_{+}) - \ell_t(y^t_{-})\bigr] s_t$

{Update} $x^{t + 1} = \Pi_{\Xcal}\Big[x^t - \frac{\alpha}{\sqrt{\sum_{i=1}^t \|g_i\|^2}} g_t\Big]$ and skip if $g_t = 0$\\
}
\caption{{\bgd} with {\adanorm} learning rate  \label{alg:adagrad-norm bgd}}
\end{algorithm}

\subsubsection{Auxiliary results}
\Cref{lem:smooth-zo} presents properties of the smoothed function $\tilde \ell$. 
Parts of these results can be found in \cite{gao2018information}. 
%A summary is provided in the lecture note \url{https://tyj518.github.io/files/lecture_notes_zo.pdf}. 

By the divergence theorem and symmetry of the unit sphere, the uniform-ball smoothing in \eqref{eq:smoothing function} satisfies
\[
  \textstyle \nabla \tilde{\ell}_t (x^t) = \frac{n}{\mu} \Ebb_{s_t \sim \Sbb_n} [\ell_t (x^t + \mu s_t) s_t] = \Ebb_{s_t \sim \Sbb_n} [g_t | x^t].
\]

\begin{lem}\label{lem:smooth-zo}
    Under \ref{A1} and \ref{A2}, the uniform-ball-smoothed function $\Tilde{\ell}_t$ defined in \eqref{eq:smoothing function} is both convex and $L$-smooth. Its gradient and function value satisfy $\|\nabla \tilde{\ell}_t(x) - \nabla \ell_t(x)\| \leq L\mu$ and $ \ell_t(x) \leq \Tilde{\ell}_t(x) \leq \ell_t(x) + \tfrac{L\mu^2}{2}$ for all $x \in \Xcal$. Moreover, the second moment of the estimator $g_t$ in \eqref{eq:gradient estimator} satisfies the bound
    \begin{equation}\label{eq:bound of second-order moment}
       \textstyle \E_{s_t\sim \Sbb_n}[\| g_t \|^2 | x^t]
       \leq 2n \|\nabla \ell_t(x^t)\|^2 + \frac{1}{2}n^2 \mu^2 L^2.
    \end{equation}
\end{lem}

\subsubsection{Proof of Theorem \ref{thm:adaptive-bgd}}
First, invoking \Cref{thm:auxi-adanorm}, we have
\begin{align*}
   \textstyle \sum_{t=1}^T \E [\inner{\nabla \Tilde{\ell}_t(x^t)}{x^t - x}] = \sum_{t=1}^T \E [\inner{\E [g_t | x^t]}{x^t - x}] \leq (\frac{D^2}{2\alpha} + \alpha)
 \E \Big[ \sqrt{\sum_{t = 1}^T \| g_t \|^2}\Big].
\end{align*}
Applying the property of dual strong convexity yields
\begin{align*}
   \textstyle  \sum_{t=1}^T \E [\Tilde{\ell}_t(x^t) - \Tilde{\ell}_t(x)] \leq{} & \textstyle (\frac{D^2}{2\alpha} + \alpha)\E \big[ \sqrt{\sum_{t=1}^T \|g_t\|^2}\big] - \frac{1}{2L} \sum_{t=1}^T \E [\|\nabla \Tilde{\ell}_t(x^t) - \nabla \Tilde{\ell}_t(x)\|^2] \\
     \leq{} & \textstyle (\frac{D^2}{2\alpha} + \alpha)\sqrt{\sum_{t=1}^T \E [ \|g_t\|^2]}- \frac{1}{2L} \sum_{t=1}^T \E [\|\nabla \Tilde{\ell}_t(x^t) - \nabla \Tilde{\ell}_t(x)\|^2] \\
     \leq{} & \textstyle (\frac{D^2}{2\alpha} + \alpha)\sqrt{\sum_{t=1}^T 2n \E [\|\nabla \ell_t(x^t)\|^2] + \frac{T n^2 L^2 \mu^2}{2}}- \frac{1}{2L} \sum_{t=1}^T \E [\|\nabla \Tilde{\ell}_t(x^t) - \nabla \Tilde{\ell}_t(x)\|^2],
\end{align*}
where we use \Cref{lem:smooth-zo} in the last step.
Choose $\alpha = \tfrac{\sqrt{2}D}{2}$. We conclude
\begin{align*}
  \textstyle\sum_{t=1}^T \E [\Tilde{\ell}_t(x^t) - \Tilde{\ell}_t(x)] \leq{} & \textstyle\sqrt{2}D \sqrt{\sum_{t=1}^T 2n \E [\|\nabla \ell_t(x^t)\|^2]} - \frac{1}{2L} \sum_{t=1}^T \E [\|\nabla \Tilde{\ell}_t(x^t) - \nabla \Tilde{\ell}_t(x)\|^2] + D nL\mu\sqrt{T} \\
  \leq{} & \textstyle 2D\sqrt{n} \sqrt{\sum_{t=1}^T \E [\|\nabla \Tilde{\ell}_t(x^t)\|^2]} - \frac{1}{2L} \sum_{t=1}^T \E [\|\nabla \Tilde{\ell}_t(x^t) - \nabla \Tilde{\ell}_t(x)\|^2] + 2D nL\mu\sqrt{T},
\end{align*}
where the last step follows from $\|\nabla \Tilde{\ell}_t (x^t) - \nabla \ell_t (x^t)\| \leq L\mu$ and $2\sqrt{n} \leq n$ for $n \geq 8$. To proceed, note that
\begin{align*}
   \textstyle \E [\|\nabla \Tilde{\ell}_t(x^t) - \nabla \Tilde{\ell}_t(x)\|^2] \ge (1-\gamma)\E [\|\nabla \Tilde{\ell}_t(x^t)\|^2] + (1 -\gamma^{-1})\|\nabla \Tilde{\ell}_t(x)\|^2, \text{~for all~} \gamma \in (0,1],
\end{align*}
which implies
\begin{align*}
   & \textstyle \sum_{t=1}^T \E [\Tilde{\ell}_t(x^t) - \Tilde{\ell}_t(x)] \\
   \leq \ &\textstyle 2D\sqrt{n} \sqrt{\sum_{t=1}^T \E [\|\nabla \Tilde{\ell}_t(x^t)\|^2]} - \frac{1-\gamma}{2L}  \sum_{t=1}^T \E [\|\nabla \Tilde{\ell}_t(x^t)\|^2] + \frac{\gamma^{-1} - 1}{2L}\sum_{t=1}^T \|\nabla \Tilde{\ell}_t(x)\|^2+ 2D nL\mu\sqrt{T}.
\end{align*}
Next, we do case analysis.

\paragraph{Case 1.} $\sum_{t=1}^T \E [\|\nabla \Tilde{\ell}_t(x^t)\|^2] \leq 64L^2 D^2 n$. Choosing $\gamma = 1$ gives
\begin{align*}
  \textstyle \sum_{t=1}^T \E [\Tilde{\ell}_t(x^t) - \Tilde{\ell}_t(x)] \leq 2D\sqrt{n}\sqrt{\sum_{t=1}^T \E [\|\nabla \Tilde{\ell}_t(x^t)\|^2]} + 2DnL\mu\sqrt{T} \le 16 LD^2 n + 2DnL\mu\sqrt{T}.  
\end{align*}

\tmtextbf{Case 2.} $\sum_{t=1}^T \E [\|\nabla \Tilde{\ell}_t(x^t)\|^2] \geq 64L^2 D^2 n$. Then taking $ \gamma = 1 - \tfrac{4LD\sqrt{n}}{\sqrt{\sum_{t=1}^T \E [\|\nabla \Tilde{\ell}_t(x^t)\|^2] }} \ge \tfrac{1}{2}$ gives
\begin{align*}
    \textstyle 2D\sqrt{n} \sqrt{\sum_{t=1}^T \E [\|\nabla \Tilde{\ell}_t(x^t)\|^2]} - \frac{1-\gamma}{2L}  \sum_{t=1}^T \E [\|\nabla \Tilde{\ell}_t(x^t)\|^2] = 0.
\end{align*}
Consequently, we deduce that
\begin{align*}
   \textstyle \sum_{t=1}^T \E [\Tilde{\ell}_t(x^t) - \Tilde{\ell}_t(x)] \le \  & \textstyle\frac{\gamma^{-1} - 1}{2L}\sum_{t=1}^T \|\nabla \Tilde{\ell}_t(x)\|^2+ 2D nL\mu\sqrt{T} \\
   = \ & \textstyle\frac{2D\sqrt{n}}{\sqrt{\sum_{t=1}^T \E [\|\nabla \Tilde{\ell}_t(x^t)\|^2] } - 4 L D \sqrt{n}}\sum_{t=1}^T \|\nabla \Tilde{\ell}_t(x)\|^2+ 2D nL\mu\sqrt{T} \\
   \le \ & \textstyle\frac{4D\sqrt{n}}{\sqrt{\sum_{t=1}^T \E [\|\nabla \Tilde{\ell}_t(x^t)\|^2] }}\sum_{t=1}^T \|\nabla \Tilde{\ell}_t(x)\|^2+ 2D nL\mu\sqrt{T},
\end{align*}
where the last inequality uses 
\begin{align*}
    \textstyle \sqrt{\sum_{t=1}^T \E [\|\nabla \Tilde{\ell}_t(x^t)\|^2] } - 4 L D \sqrt{n} \ge \tfrac{1}{2}\sqrt{\sum_{t=1}^T \E [\|\nabla \Tilde{\ell}_t(x^t)\|^2] }
\end{align*}
since $\sqrt{\sum_{t=1}^T \E [\|\nabla \Tilde{\ell}_t(x^t)\|^2] } \ge 8LD\sqrt{n}$. Together with $\textstyle \sum_{t=1}^T \E [\Tilde{\ell}_t(x^t) - \Tilde{\ell}_t(x)] \leq 2D\sqrt{n} \sqrt{\sum_{t=1}^T \E [\|\nabla \Tilde{\ell}_t(x^t)\|^2]} + 2D nL\mu\sqrt{T}$, applying
$\min \{ x, \tfrac{a}{x} \} \leq \sqrt{a}$ gives
\begin{align*}
   \textstyle \sum_{t=1}^T \E [\Tilde{\ell}_t(x^t) - \Tilde{\ell}_t(x)] \le{} & \textstyle 2D\sqrt{n} \min\{\sqrt{\sum_{t=1}^T \E [\|\nabla \Tilde{\ell}_t(x^t)\|^2]}, \tfrac{2\sum_{t=1}^T \|\nabla \Tilde{\ell}_t(x)\|^2}{\sqrt{\sum_{t=1}^T \E [\|\nabla \Tilde{\ell}_t(x^t)\|^2]}}\} + 2DnL\mu\sqrt{T} \\
   \le \ & \textstyle 2\sqrt{2}D\sqrt{n}\sqrt{\sum_{t=1}^T \|\nabla \Tilde{\ell}_t(x)\|^2}  + 2DnL\mu\sqrt{T}.
\end{align*}
Now, putting the two cases together gives
\begin{align*}
     \textstyle \sum_{t=1}^T \E [\Tilde{\ell}_t(x^t) - \Tilde{\ell}_t(x)] \le \max\{16LD^2n, 2\sqrt{2}D\sqrt{n}\sqrt{\sum_{t=1}^T \|\nabla \Tilde{\ell}_t(x)\|^2}\} + 2DnL\mu\sqrt{T}.
\end{align*}
Finally, using the inequalities $\ell_t(x) \leq \Tilde{\ell}_t(x) \leq \ell_t(x) + \tfrac{L\mu^2}{2}$ and $\|\nabla \Tilde{\ell}_t(x) - \nabla \ell_t(x)\| \leq L\mu$ for any $x \in \Xcal$, we have
\begin{align*}
     \textstyle\E [\rho_T(x)] \le \max\{16LD^2n, 4D\sqrt{n G_T(x)}\} + 2DL\mu\sqrt{T}(n+2\sqrt{n}) + \tfrac{TL\mu^2}{2}.
\end{align*}
Recall that $2\sqrt{n} \le n$ since $n \ge 8$. Substituting $\mu = \tfrac{D}{n\sqrt{T}}$ gives $\textstyle  \E [\rho_T(x)] \le \max\{16LD^2n, 4D\sqrt{n G_T(x)}\} + 5LD^2$. The proof is complete.

\section{Proof of results in Section \ref{sec:implication}}

\subsection{Proof of Lemma \ref{lem:holder-gstar-conversion}}

For simplicity, let $p \assign \frac{1 + \nu}{\nu}$ and $d_t \assign a_t - b_t$. By the variational representation of the Euclidean norm,
\begin{align*}
  \textstyle \alpha \sqrt{\sum_{t = 1}^T \| b_t + d_t \|^2}
  ={} & \textstyle \max_{\sum_{t = 1}^T \| u_t \|^2 \leq \alpha^2} \sum_{t = 1}^T \langle u_t, b_t + d_t \rangle \\
  \leq{} & \textstyle \alpha \sqrt{\sum_{t = 1}^T \| b_t \|^2} + \max_{\sum_{t = 1}^T \| u_t \|^2 \leq \alpha^2} \sum_{t = 1}^T \langle u_t, d_t \rangle.
\end{align*}
It follows that
\begin{align*}
  & \textstyle \alpha \sqrt{\sum_{t = 1}^T \| a_t \|^2} - \beta \sum_{t = 1}^T \| a_t - b_t \|^p \\
  \leq{} & \textstyle \alpha \sqrt{\sum_{t = 1}^T \| b_t \|^2} + \max_{\sum_{t = 1}^T \| u_t \|^2 \leq \alpha^2} \sum_{t = 1}^T \sup_{v_t} \{ \langle u_t, v_t \rangle - \beta \| v_t \|^p \}.
\end{align*}
For every $t$, the supremum is
\[
  \textstyle \sup_{v_t} \{ \langle u_t, v_t \rangle - \beta \| v_t \|^p \} = (p - 1) p^{-\frac{p}{p - 1}} \beta^{-\frac{1}{p - 1}} \| u_t \|^{\frac{p}{p - 1}} = \frac{\nu^\nu}{(1 + \nu)^{1 + \nu}} \beta^{-\nu} \| u_t \|^{1 + \nu}.
\]
Moreover, H\"{o}lder's inequality gives
\[
  \textstyle \sum_{t = 1}^T \| u_t \|^{1 + \nu} \leq T^{\frac{1 - \nu}{2}} (\sum_{t = 1}^T \| u_t \|^2)^{\frac{1 + \nu}{2}}.
\]
Combining the last three displays and using $\sum_{t = 1}^T \| u_t \|^2 \leq \alpha^2$ proves the result.

\subsection{Proof of Theorem \ref{thm:holder-gstar}}

By \ref{A3} and the assumed linearized-regret guarantee,
\begin{align*}
  \rho_T (x)
  \leq{} & \textstyle \sum_{t = 1}^T \langle \nabla \ell_t (x^t), x^t - x \rangle - \frac{\nu}{1 + \nu} L^{-1 / \nu} \sum_{t = 1}^T \| \nabla \ell_t (x^t) - \nabla \ell_t (x) \|^{\frac{1 + \nu}{\nu}} \\
  \leq{} & \textstyle \delta (x) \sqrt{\sum_{t = 1}^T \| \nabla \ell_t (x^t) \|^2} - \frac{\nu}{1 + \nu} L^{-1 / \nu} \sum_{t = 1}^T \| \nabla \ell_t (x^t) - \nabla \ell_t (x) \|^{\frac{1 + \nu}{\nu}}.
\end{align*}
Apply \Cref{lem:holder-gstar-conversion} with $a_t = \nabla \ell_t (x^t)$, $b_t = \nabla \ell_t (x)$, $\alpha = \delta (x)$, and $\beta = \frac{\nu}{1 + \nu} L^{-1 / \nu}$. This gives
\begin{align*}
  \rho_T (x)
  \leq{} & \textstyle \delta (x) \sqrt{G_T (x)} + \frac{\nu^\nu}{(1 + \nu)^{1 + \nu}} \frac{\delta (x)^{1 + \nu}}{(\frac{\nu}{1 + \nu} L^{-1 / \nu})^\nu} T^{\frac{1 - \nu}{2}} \\
  ={} & \textstyle \delta (x) \sqrt{G_T (x)} + \frac{L}{1 + \nu} \delta (x)^{1 + \nu} T^{\frac{1 - \nu}{2}},
\end{align*}
which proves the result.

\subsection{Proof of Corollary \ref{coro:holder-adanorm}}

By \Cref{thm:auxi-adanorm}, the linearized regret of {\adanorm} satisfies
\[
  \textstyle \sum_{t = 1}^T \langle \nabla \ell_t (x^t), x^t - x \rangle \leq (\alpha + \frac{D^2}{2 \alpha}) \sqrt{\sum_{t = 1}^T \| \nabla \ell_t (x^t) \|^2}
\]
for any $x \in \Xcal$. Taking $\alpha = \frac{\sqrt{2}D}{2}$ gives $\alpha + \frac{D^2}{2 \alpha} = \sqrt{2}D$. The result follows from \Cref{thm:holder-gstar} with $\delta (x) = \sqrt{2}D$.

\subsection{Proof of Corollary \ref{coro:holder-gnu-upper}}

Since $\frac{1 + \nu}{\nu} \geq 2$, H\"{o}lder's inequality gives
\[
  \textstyle \sqrt{G_T (x)} = (\sum_{t = 1}^T \| \nabla \ell_t (x) \|^2)^{\frac{1}{2}} \leq T^{\frac{1 - \nu}{2 (1 + \nu)}} (\sum_{t = 1}^T \| \nabla \ell_t (x) \|^{\frac{1 + \nu}{\nu}})^{\frac{\nu}{1 + \nu}} = T^{\frac{1 - \nu}{2 (1 + \nu)}} G_{\nu,T} (x)^{\frac{\nu}{1 + \nu}}.
\]
Substituting this inequality into \Cref{thm:holder-gstar} proves the result.

\subsection{Proof of Lemma \ref{lem:log-sequence}}

Let $A \assign \sum_{t = 1}^T \| a_t \|^2$, $B \assign \sum_{t = 1}^T \| b_t \|^2$, and $U \assign \sum_{t = 1}^T \| a_t - b_t \|^2$. Since $A \leq 2 B + 2 U$, the left-hand side in the lemma is at most
\[
  \textstyle \alpha \log (1 + 2 \gamma B + 2 \gamma U) - \beta U.
\]
Set $c \assign 1 + 2 \gamma B$ and $h (u) \assign \alpha \log (c + 2 \gamma u) - \beta u$ for $u \geq 0$. If $c \geq \frac{2 \alpha \gamma}{\beta}$, then $h$ is non-increasing and
\[
  \textstyle h (U) \leq h (0) = \alpha \log (1 + 2 \gamma B).
\]
Otherwise, $h$ is maximized at $u^{\star} = \frac{\alpha}{\beta} - \frac{c}{2 \gamma} > 0$, and $  \textstyle h (U) \leq h (u^{\star}) = \alpha \log (\frac{2 \alpha \gamma}{\beta}) - \alpha + \frac{\beta c}{2 \gamma} \leq \alpha \log (\frac{2 \alpha \gamma}{\beta})$.
Taking the larger of the two bounds proves the result.

\subsection{Proof of Theorem \ref{thm:strongly-convex-gstar}}

Let $g_t \assign \nabla \ell_t (x^t)$, $b_t \assign \nabla \ell_t (x)$, and $S_t \assign \sum_{i = 1}^t \| g_i \|^2$. By \ref{A1} and \ref{A4}, the Bregman divergence of each loss satisfies
\[
  \textstyle \ell_t (x) - \ell_t (x^t) - \langle g_t, x - x^t \rangle \geq \frac{\mu}{4} \| x^t - x \|^2 + \frac{1}{4 L} \| g_t - b_t \|^2.
\]
Therefore,
\begin{equation} \label{eqn:strong-gstar-proof-1}
  \textstyle \rho_T (x) \leq \sum_{t = 1}^T [\langle g_t, x^t - x \rangle - \frac{\mu}{4} \| x^t - x \|^2] - \frac{1}{4 L} \sum_{t = 1}^T \| g_t - b_t \|^2.
\end{equation}
Define $\Delta \assign \frac{8 L M^2}{\mu}$, so the learning rate in \Cref{thm:strongly-convex-gstar} is $\eta_t = \frac{2 M^2 / \mu}{\Delta + S_t}$. Standard OCO analysis gives $\textstyle \langle g_t, x^t - x \rangle \leq \frac{\| x^t - x \|^2 - \| x^{t + 1} - x \|^2}{2 \eta_t} + \frac{\eta_t}{2} \| g_t \|^2$.
For $t \geq 2$, telescoping gives $\textstyle \frac{1}{2 \eta_t} - \frac{1}{2 \eta_{t - 1}} = \frac{\mu \| g_t \|^2}{4 M^2} \leq \frac{\mu}{4}$.
The initial coefficient satisfies $\frac{1}{2 \eta_1} - \frac{\mu}{4} = \frac{\mu (\Delta + \| g_1 \|^2 - M^2)}{4 M^2} \leq \frac{\mu \Delta}{4 M^2}$.
Using \ref{A2}, dropping the non-positive final potential, and telescoping yields 
\begin{align*}
  \textstyle \sum_{t = 1}^T [\langle g_t, x^t - x \rangle - \frac{\mu}{4} \| x^t - x \|^2]
  \leq{} & \textstyle \frac{\mu \Delta D^2}{4 M^2} + \frac{M^2}{\mu} \sum_{t = 1}^T \frac{\| g_t \|^2}{\Delta + S_t} \\
  \leq{} & \textstyle \frac{\mu \Delta D^2}{4 M^2} + \frac{M^2}{\mu} \log (1 + \frac{S_T}{\Delta}) \\
  ={} & \textstyle 2 L D^2 + \frac{M^2}{\mu} \log (1 + \frac{\mu}{8 L M^2} \sum_{t = 1}^T \| g_t \|^2).
\end{align*}
Substitute this bound into \eqref{eqn:strong-gstar-proof-1} and apply \Cref{lem:log-sequence} with $a_t = g_t, b_t = \nabla \ell_t (x), \alpha = \frac{M^2}{\mu}, \beta = \frac{1}{4 L},  \gamma = \frac{\mu}{8 L M^2}.$
Here $\frac{2 \alpha \gamma}{\beta} = 1$, with $\sum_{t = 1}^T \| b_t \|^2 = G_T (x)$. Hence
\[
  \textstyle \rho_T (x) \leq 2 L D^2 + \frac{M^2}{\mu} \log (1 + \frac{\mu G_T (x)}{4 L M^2}),
\]
which proves the result.

\subsection{Proof of Corollary \ref{coro:stochastic-opt}}
We first prove the result for {\ogd}. The standard {\ogd} analysis under \ref{A1} gives, for any $x \in \mathcal{X}$ and $\eta \in (0, \tfrac{1}{L})$,
\[
  \textstyle \sum_{t = 1}^T \ell_t (x^t) - \ell_t (x) \leq \tfrac{1}{2 \eta} \| x^1 - x \|^2 + \tfrac{\eta}{2(1 - \eta L)} \sum_{t = 1}^T \| \nabla \ell_t (x) \|^2.
\]
Let $x = x^\star$ and $\ell_t(x) = f(x, \xi^t)$. We have
\begin{align*}
   \textstyle \sum_{t = 1}^T [f(x^t, \xi^t) - f(x^\star, \xi^t)] \leq \tfrac{1}{2 \eta} \| x^1 -
     x^\star \|^2 + \tfrac{\eta}{2(1 - \eta L)}  \sum_{t = 1}^T \| \nabla f(x^\star, \xi^t) \|^2.
\end{align*}
Note that $\E_{\xi^t} [f(x^t, \xi^t) | x^t] = f(x^t)$ and $\E_{\xi^t} [f(x^\star, \xi^t) | x^t] = f(x^\star)$. Taking expectation and using convexity,
\begin{align*}
T \E [f(\bar{x}^T) - f(x^\star)]\leq{} &
  \textstyle \E \big[\sum_{t = 1}^T [f(x^t) - f(x^\star)]\big] \\
  \leq{} & \textstyle \tfrac{1}{2 \eta} \| x^1 -
     x^\star \|^2 + \tfrac{\eta}{2(1 - \eta L)}  \sum_{t = 1}^T \E [ \| \nabla f(x^\star, \xi^t) \|^2 ] \\
     ={} &\tfrac{1}{2 \eta} \| x^1 -
     x^\star \|^2 + \tfrac{\eta}{2(1 - \eta L)} T (\sigma_g^\star)^2.
\end{align*}
By a similar online-to-batch argument, the corresponding results for {\adanorm} and {\adaftrl} can be proved by noticing that $\E [\sqrt{\sum_{t = 1}^T \| \nabla f (x^\star, \xi^t) \|^2}] \leq \sqrt{\E [\sum_{t = 1}^T \| \nabla f (x^\star, \xi^t) \|^2]} = \sqrt{T} \sigma_g^\star$.

\subsection{Proof of Theorem \ref{thm:adanorm-opt}}

We show the result by establishing a regret version of the result with
$G_T^{\star} = 0$. i.e. $\nabla \ell_t (x^{\star}) = 0$ for all $t$. Recall the update of {\adanorm} for optimization problem $x^{t + 1} = x^t - \eta_t \nabla \ell_t
(x^t)$, where $\eta_t = \tfrac{\alpha}{\sqrt{\sum_{j = 1}^t \| \nabla \ell_t (x^t)
\|^2}}$. 

We deduce that
\begin{align}
  \| x^{t + 1} - x \|^2={} & \| x^t - \eta_t \nabla \ell_t (x^t) - x \|^2 \nonumber\\
  ={} & \| x^t - x \|^2 - 2 \eta_t \langle \nabla \ell_t (x^t), x^t - x \rangle
  + \eta_t^2 \| \nabla \ell_t (x^t) \|^2 \nonumber\\
  \leq{} & \| x^t - x \|^2 - 2 \eta_t [\ell_t (x^t) - \ell_t (x)] +
  \tfrac{\eta_t^2}{1 - \eta_t L} \| \nabla \ell_t (x) \|^2 + ( \eta^2_t -
  \tfrac{\eta_t}{L} ) \| \nabla \ell_t (x^t) - \tfrac{1}{1 -
  \eta_t L} \nabla \ell_t (x) \|^2 \nonumber
\end{align}
Taking $x = x^{\star}$, we have $\ell_t (x^t) - \ell_t (x^{\star}) \geq 0$ and
that $\nabla \ell_t (x^{\star}) = 0$, giving
\begin{align}
 \| x^{t + 1} - x^{\star} \|^2 
  \leq{} & \textstyle \| x^1 - x^{\star} \|^2 + \sum_{j = 1}^t ( \eta^2_j -
  \tfrac{\eta_j}{L} ) \| \nabla \ell_j (x^j) \|^2 \nonumber\\
  \leq{} & \textstyle\| x^1 - x^{\star} \|^2 + ( \eta^2_1 - \frac{\eta_1}{L} )
  \| \nabla \ell_1 (x^1) \|^2 + \sum_{j \in \{ k : k \geq 2, \eta_k \geq
  \frac{1}{L} \}} ( \eta^2_j - \tfrac{\eta_j}{L} ) \| \nabla
  \ell_j (x^j) \|^2 \nonumber\\
  ={} & \textstyle \| x^1 - x^{\star} \|^2 + ( \alpha^2 - \tfrac{\alpha}{L} \| \nabla
  \ell_1 (x^1) \| ) + \sum_{j \in \{ k : k \geq 2, \eta_k \geq
  \frac{1}{L} \}} ( \eta^2_j - \tfrac{\eta_j}{L} ) \| \nabla
  \ell_j (x^j) \|^2 \nonumber
\end{align}

since $\eta^2 - \tfrac{\eta}{L} \leq 0$ for $\eta \leq \tfrac{1}{L}$. Next we
bound $\sum_{j \in \{ k : k \geq 2, \eta_k \geq \tfrac{1}{L} \}}
( \eta^2_j - \tfrac{\eta_j}{L} ) \| \nabla \ell_j (x^j) \|^2$ by
taking $a_0 = \| \nabla \ell_1 (x^1) \|^2$, $a_j = \| \nabla \ell_{j + 1}
(x^{j + 1}) \|^2$ and $g (u) = \tfrac{\alpha^2}{u} - \frac{\alpha}{L
\sqrt{u}}$ in \Cref{lem:adanorm}:
\begin{align}
\textstyle  \sum_{j \in \{ k : k \geq 2, \eta_k \geq{} \frac{1}{L} \}} (
  \eta^2_j - \tfrac{\eta_j}{L} ) \| \nabla \ell_j (x^j) \|^2 \leq{} & \textstyle
  \int_{\| \nabla \ell_1 (x^1) \|^2}^{(\alpha L)^2} g (u) ~\mathd u \nonumber\\
  ={} & 2 \alpha^2 \log \tfrac{\alpha L}{\| \nabla \ell_1 (x^1) \|} - \tfrac{2
  \alpha}{L} [\alpha L - \| \nabla \ell_1 (x^1) \|] \nonumber\\
  ={} & 2 \alpha^2 \log \tfrac{\alpha L}{\| \nabla \ell_1 (x^1) \|} - 2 \alpha^2
  + \tfrac{2 \alpha}{L} \| \nabla \ell_1 (x^1) \| \nonumber
\end{align}
Putting things together, we have
\begin{align}
  \| x^{t + 1} - x^{\star} \|^2 \leq{} & \| x^1 - x^{\star} \|^2 + \alpha^2 -
  \tfrac{\alpha}{L} \| \nabla \ell_1 (x^1) \| + 2 \alpha^2 \log \tfrac{\alpha
  L}{\| \nabla \ell_1 (x^1) \|} - 2 \alpha^2 + \tfrac{2 \alpha}{L} \| \nabla
  \ell_1 (x^1) \| \nonumber\\
  ={} & \| x^1 - x^{\star} \|^2 + \tfrac{\alpha}{L} \| \nabla \ell_1 (x^1) \| +
  2 \alpha^2 \log \tfrac{\alpha L}{\sqrt{\mathe} \| \nabla \ell_1 (x^1) \|}
 =:{} \hat{D} \nonumber
\end{align}
for all $t \geq 1$. Hence, the standard convergence analysis of {\adanorm} with $D
= \hat{D}$ applies. By \Cref{thm:auxi-adanorm},
\begin{align}
 \textstyle \sum_{t = 1}^T \ell_t (x^t) - \ell_t (x^{\star}) \leq{} & (
\textstyle  \tfrac{\hat{D}}{2 \alpha} + \alpha ) \sqrt{\sum_{j = 1}^t \| \nabla
  \ell_t (x^t) \|^2} - \tfrac{1}{2 L} \sum_{j = 1}^t \| \nabla \ell_t (x^t)
  \|^2 \leq \tfrac{L}{2} ( \tfrac{\hat{D}}{2 \alpha} + \alpha )^2 .
  \nonumber
\end{align}
Plugging in the definition of $\hat{D}$ gives a convergence rate of
\begin{align}
 \textstyle \sum_{t = 1}^T \ell_t (x^t) - \ell_t (x^{\star}) \leq{} & \textstyle \tfrac{L}{2} \big(
  \tfrac{\| x^1 - x^{\star} \|^2}{2 \alpha} + \alpha \log (
  \tfrac{\sqrt{\mathe} \alpha L}{\| \nabla \ell_t (x^1) \|} ) +
  \tfrac{1}{2 L} \| \nabla \ell_t (x^1) \| \big)^2 \nonumber
\end{align}
and an online-to-batch argument completes the proof.

\subsection{Proof of Proposition \ref{prop:hardness}}

The construction modifies the standard OCO lower bound. Consider a 1D OCO
problem with
\[ \ell_t (x) = \tfrac{1 + \sigma_t}{2} g (x) + \tfrac{1 - \sigma_t}{2} g (- x) \]
where $g (x) = \bigg\{ \begin{array}{ll}
  x + \tfrac{3}{2}, & x \geq - 1\\
  \tfrac{1}{2} [x + 2]_+^2, & \text{else}
\end{array} $ and $\sigma_t \in \{ - 1, 1 \}$ is a Rademacher random
variable. It is easy to check that $g$ is $L = 1$-smooth, convex, with $\inf g
= 0$. Moreover, for $\mathcal{X}= [- \delta, \delta] \subseteq [- 1, 1]$, we
have $\ell_t (x) = \sigma_t x + \tfrac{3}{2}$ and therefore, the regret in terms
of $\{ \ell_t \}_{t = 1}^T$ coincides with that of the 1D OCO lower bound
since for $x^t, x \in \mathcal{X}$, we have
\begin{align}
  \rho_T (x) ={} & \textstyle \sum_{t = 1}^T \ell_t (x^t) - \sum_{t = 1}^T \ell_t (x)
  \nonumber\\
  ={} & \textstyle  \sum_{t = 1}^T [ \sigma_t x^t + \tfrac{3}{2} ] - \sum_{t =
  1}^T [ \sigma_t x + \tfrac{3}{2} ] \nonumber\\
  ={} & \textstyle  \sum_{t = 1}^T \sigma_t (x^t - x). \nonumber
\end{align}

Furthermore, we can upperbound $L^{\mathcal{X}}_T (x)$ by
\[ \textstyle L^{\mathcal{X}}_T (x) = \sum_{t = 1}^T \ell_t (x) - \min_{u \in
   \mathcal{X}} \ell_t (u) = \sum_{t = 1}^T \sigma_t x - \min_{u \in [-
   \delta, \delta]} \sigma_t u \leq 2 \delta T. \]
Suppose an $\mathcal{O} ( L D^2 + D\sqrt{L L^{\mathcal{X}}_T (x^{\star})}
)$ upper bound holds, then with $D = 2 \delta$, it would imply
\[ \rho_T (x) \leq \mathcal{O} ( 4 \delta^2 + 2 \delta \sqrt{2 \delta T}
   ) =\mathcal{O} ( \delta^2 + \delta^{3 / 2} \sqrt{T} ). \]
However, the lower bound of OCO implies $\rho_T (x) \geq \Omega ( D
\sqrt{T} ) = \Omega ( \delta \sqrt{T} )$, and taking $\delta
= T^{- \beta}, \beta \in ( 0, \tfrac{1}{3} ]$ leads to an upper
bound of $\mathcal{O} ( T^{\frac{1 - 3 \beta}{2}} )$ and a lower
bound of $\Omega ( T^{\frac{1}{2} - \beta} )$ that contradict each
other. Dividing the upper and lower bounds completes the proof.

\subsection{Proof of Proposition \ref{prop:smooth}}
We have $G^{\mathcal{X}}_T (x) \leq 2 \gamma [ \sum_{t = 1}^T \ell^{1 /
  \gamma}_t (x) - \inf \ell^{1 / \gamma}_t ]$  due to smoothness of $\ell^{1/\gamma}(x)$. The right-hand side can be further bounded by
\begin{align*}
	\textstyle \sum_{t = 1}^T \ell^{1 /
  \gamma}_t (x) - \inf \ell^{1 / \gamma}_t ={} & \textstyle \sum_{t = 1}^T \ell^{1 /
  \gamma}_t (x) - \min_{u \in \Xcal} \ell_t(u) \\
  \leq{} & \textstyle  \sum_{t = 1}^T \ell^{1 /
  \gamma}_t (x) - \min_{u \in \Xcal} \ell_t(u) = L^{\mathcal{X}}_T (x),
\end{align*}
  where the equality uses the fact that $\inf \ell^{1 / \gamma}_t = \min_{u \in \Xcal} \ell_t(u)$ and
   the last inequality uses  $\ell(x) \leq \ell^{1/\gamma}(x)$ from \eqref{eqn:envelop-approx}. This completes the proof.
   
\subsection{Hardness of achieving an $\Ocal(D\sqrt{G^\Xcal_T(x)})$ regret upper bound}

We show that an $\mathcal{O} ( L D^2 + D
\sqrt{G_T^{\mathcal{X}} (x)} )$ upper bound is also generally not
achievable.

\begin{prop}
  Let $\gamma = \Omega (1)$ be some constant. For any $\tau \in [ 0,
  \tfrac{1}{2} )$, there exists a loss sequence $\{ \ell_t \}_{t = 1}^T$
  satisfying \ref{A1}, \ref{A2} such that for any online algorithm
  and $x \in \mathcal{X}$, $\tfrac{\rho_T (x)}{L D^2 + D
  \sqrt{G_T^{\mathcal{X}} (x)}} = \Omega (T^{\tau})$.
\end{prop}

\begin{proof}
	
Consider the lower bound instance with $\mathcal{X}= [- \delta, \delta]$, $D =
2 \delta$, and $\ell_t (x) = g_t x$, where $g_t = \sigma_t$ is a Rademacher
random variable. Each $\ell_t$ is linear and thus 0-smooth. Then
\[ \ell_t^{1 / \gamma} (x) = \min_{z \in [- \delta, \delta]}  \{ g_t z +
   \tfrac{\gamma}{2} (z - x)^2\}, \]
and for $x \in \mathcal{X}$, we have
\[ \| \nabla \ell_t^{1 / \gamma} (x) \|^2 = \gamma^2 | \Pi_{[- \delta,
   \delta]} [ x - \tfrac{1}{\gamma} g_t ] - x |^2 \leq 4
   \gamma^2 \delta^2. \]
Hence, $G_T^{\mathcal{X}} (x) \leq 4 \gamma^2 \delta^2 T$. The lower
bound argument gives $\rho_T (x) \geq \Omega ( \delta \sqrt{T} )$,
while the desired $\mathcal{O} ( L D^2 + D \sqrt{G_T^{\mathcal{X}} (x)}
)$ upper bound with $L = 0$ yields
\[ \rho_T (x) =\mathcal{O} ( \delta \gamma \sqrt{4 \delta^2 T} ) =
   \gamma \delta^2 \sqrt{T} . \]
Taking $\delta = T^{- \beta}$, we have lower bound of $\Omega (
T^{\frac{1}{2} - \beta} )$ and an upper bound of $\mathcal{O} (
\gamma T^{\frac{1}{2} - 2 \beta} )$ that contradict each other.

\end{proof}